\documentclass[sigconf]{acmart}
\renewcommand\footnotetextcopyrightpermission[1]{} % removes footnote with conference information in first column
\usepackage[ruled]{algorithm2e} % For algorithms
\usepackage[export]{adjustbox}

\usepackage[font=small]{caption}
\usepackage[labelformat = empty,position=top]{subcaption} 

\AtBeginDocument{%
  \providecommand\BibTeX{{%
    \normalfont B\kern-0.5em{\scshape i\kern-0.25em b}\kern-0.8em\TeX}}}

\setcopyright{rightsretained}

\begin{document}

\title{Concept-based model explanations for Electronic Health Records}

\author{Diana Mincu}
\affiliation{%
  \institution{Google Research}
  \city{London}
  \country{UK}
}

\author{Eric Loreaux}
\affiliation{%
  \institution{Google Health}
  \city{Palo Alto}
  \country{CA, USA}
}

\author{Shaobo Hou}
\affiliation{%
  \institution{DeepMind}
  \city{London}
  \country{UK}
}

\author{Sebastien Baur}
\affiliation{%
  \institution{Google Health}
  \city{London}
  \country{UK}
}

\author{Ivan Protsyuk}
\affiliation{%
  \institution{Google Health}
  \city{London}
  \country{UK}
}

\author{Martin Seneviratne}
\affiliation{%
  \institution{Google Health}
  \city{London}
  \country{UK}
}

\author{Anne Mottram}
\affiliation{%
  \institution{DeepMind}
  \city{London}
  \country{UK}
}

\author{Nenad Tomasev}
\affiliation{%
  \institution{Deepmind}
  \city{London}
  \country{UK}
}

\author{Alan Karthikesalingam}
\affiliation{%
  \institution{Google Health}
  \city{London}
  \country{UK}
}

\author{Jessica Schrouff}
\authornote{Corresponding author.}
\affiliation{%
  \institution{Google Research}
  \city{London}
  \country{UK}
}
\email{schrouff@google.com}

% \copyrightyear{2021}
% \acmYear{2021}
\acmConference[ACM CHIL '21]{ACM Conference on Health, Inference, and Learning}{April 8--10, 2021}{Virtual Event, USA}
% \acmBooktitle{ACM Conference on Health, Inference, and Learning (ACM CHIL '21), April 8--10, 2021, Virtual Event, USA}\acmDOI{10.1145/3450439.3451858}
% \acmISBN{978-1-4503-8359-2/21/04}

% <concept_id>10010147.10010178</concept_id>
% <concept_desc>Computing methodologies~Artificial intelligence</concept_desc>
% <concept_id>10010147.10010257</concept_id>
% <concept_desc>Computing methodologies~Machine learning</concept_desc>
% <concept_id>10010147.10010257.10010293.10010294</concept_id>
% <concept_desc>Computing methodologies~Neural networks</concept_desc>
% <concept_significance>300</concept_significance>

% \ccsdesc[500]{Computing methodologies~Artificial intelligence}
% \ccsdesc[500]{Computing methodologies~Machine learning}
% \ccsdesc[300]{Computing methodologies~Neural networks}


\renewcommand{\shortauthors}{Mincu et al.}

\begin{abstract}
Recurrent Neural Networks (RNNs) are often used for sequential modeling of adverse outcomes in electronic health records (EHRs) due to their ability to encode past clinical states. These deep, recurrent architectures have displayed increased performance compared to other modeling approaches in a number of tasks, fueling the interest in deploying deep models in clinical settings. One of the key elements in ensuring safe model deployment and building user trust is model explainability. Testing with Concept Activation Vectors (TCAV) has recently been introduced as a way of providing human-understandable explanations by comparing high-level concepts to the network’s gradients. While the technique has shown promising results in real-world imaging applications, it has not been applied to structured temporal inputs. To enable an application of TCAV to sequential predictions in the EHR, we propose an extension of the method to time series data. We evaluate the proposed approach on an open EHR benchmark from the intensive care unit, as well as synthetic data where we are able to better isolate individual effects.
\end{abstract}

\maketitle

\section{Introduction}
\label{introduction}
Wider availability of Electronic Health Records (EHR) has led to an increase in machine learning applications for clinical diagnosis and prognosis ~\citep[e.g.,][]{Ambrosino1995, Caruana2015}. Larger de-identified datasets and public benchmarks have fueled the application of increasingly complex techniques such as recurrent neural networks (RNNs) to predict adverse clinical events~\citep[e.g.,][]{Lipton2016, Xiao2018, Shickel2019, Futoma, Tomasev2019}. RNNs can operate over a sequence of health information, iteratively combining input data with internal memory states to generate new states, making them suitable for continuous clinical predictions. While these memory-storing networks allow for accurate and dynamic predictions, it is often difficult to examine the mechanism by which clinical information is being translated into outputs. In healthcare, as in other fields in which trust is paramount, it is not sufficient to show state of the art discriminative performance; clinicians have also deemed it critical that models provide local and global \textit{explanations} for their behavior ~\citep{Tonekaboni2019}.

Multiple approaches have been proposed to provide explanations for machine learning models applied to EHR data ~\citep[see][for a review]{Payrovnaziri2020}, with a focus on attention-based methods when the architecture relies on RNNs ~\cite[e.g.,][]{Choi2016, Sha2017,Shickel2019}. Typically, those interpretability techniques ranks the input features based on their attention scores. However, single feature rankings might not highlight clinical states that encompass multiple input features (e.g. ``infection'') in an intuitive manner. To address this issue of human understandability, \citet{Panigutti2020} use an ontology of diagnoses to provide insights across single features. This approach, however, relies on diagnoses which are typically recorded at the end of an admission and is therefore not suitable to identify temporal changes across features that reflect a clinical concept, nor is it able to provide continuous predictions.

On the other hand, human-understandable explanations have been successfully developed for computer vision applications: Testing with Concept Activation Vectors ~\citep[TCAV,][]{Kim2017} relies on human-understandable ``concepts'' to derive model explanations. Practitioners or end users can select examples from the data that embody intuitive concepts (e.g. ``pointy ears'' or ``stripes''), and these examples are then used to map concepts to the model's activation space in the form of concept activation vectors (CAVs). CAVs can then be used to provide global explanations, as well as assess the presence or absence of a concept in local examples.

In this work, we define ``clinical concepts'' from temporal EHR input features to improve the human-understandability of post-hoc explanations of continuous clinical predictions. Our approach leverages  TCAV ~\citep{Kim2017} and can be applied to previously trained models without restrictions on model inputs or RNN architecture. Our contributions are as follows:
\begin{itemize}
    \item We extend the TCAV approach to the time series setting by defining metrics assessing (1) whether the model encodes the concept, (2) whether the concept is ``present'' in examples, and (3) whether a concept influences the model's predictions.
    \item We design a synthetic time series dataset to evaluate (concept-based) attribution methods and demonstrate that the proposed technique is faithful.
    \item We propose a framework to define human-understandable concepts in EHR and illustrate it using the de-identified MIMIC-III benchmark dataset ~\citep{Johnson2016}.
\end{itemize}

\section{Methods}

\paragraph{Notation:} We consider a set of multivariate time series $\mathcal{X} := \big({x}_{i, t, d}\big)_{i \leq N, t \leq T_i, d \leq D}$, where $x_{i,t,d} \in \mathbb{R}$, $N$ is the number of time series (i.e. patients), $D$ is the number of features per time step and $T_i$ the number of time steps for patient $i$. We define $\textbf{x}_{d}$ as the time series for feature $d \in \{1,\dots,D\}$ for a single example. The label $\textbf{y} \in \{0, 1\}^{N \times T}$ exists for all examples and all time steps. We train a recurrent neural network $F : {\mathcal{X}} \rightarrow [0,1]^T$ with $L$ layers. For a given layer $1\leq l \leq L$ and time step $1\leq t \leq T$, we can write the predicted output of $F$ as $F_t(\vec{x}) := h(f_l(\vec{x}_{1:t}))$ where $f_l(\vec{x}_{1:t})$ is the activation vector at the $l$-th layer after $t$ time steps, further referred to as $\vec{a}_{t,l}$ and $h$ represents the operations in layers $l \dots L$. Please note that we consider binary classification settings, but the approach extends to multi-class predictions.

\subsection{Concept-based explanations over time}
\label{sec:temporal_cavs}
In this section, we extend TCAV \citep{Kim2017} to account for the temporal dimension. TCAV relies on two main steps: (1) Building a concept activation vector (CAV) for each concept, and (2) assessing how the concept influences the model's decision. 

\paragraph{Building a CAV:} To build a CAV, \citet{Kim2017} sample positive and negative examples for a concept, record their activations $\vec{a}_{l}$ at each layer $l$ of the network and build a linear classifier (e.g. logistic regression) distinguishing between activations related to positive and negative samples. To extend this approach to timeseries, we identify a `time window of interest' $[t_{start}, t_{end}]$ that reflects a trajectory corresponding to a concept, i.e. during which some features or feature changes are present. We define a `control' group as a set of trajectories in which the concept does not manifest. We then collect the model's activations from $t=t_{start}$ to the end of the window $t_{end}$ for both groups, and training data for CAV learning is defined based on three different strategies:
\begin{itemize}
    \item CAV$_{t_{end}}$: we record the model's activations in each layer at $t_{end}$. This reflects the assumption that the trajectory can be represented by its end point.
    \item CAV$_{t_{start} : t_{end}}$: we record the model's activations at each time step between $t_{start}$ and $t_{end}$, using them as samples in the linear classifier. This approach hypothesizes that each time step in the trajectory represents a key component of the concept pattern.
    \item  CAV$_{t_{end} - t_{start}}$: we record the model's activations at $t_{start}$ and at $t_{end}$ and use their difference to train the CAV. In this case, we assume that changes in activations represent the concept of interest.
\end{itemize}

A concept is considered as ``encoded'' in the model if the linear model performs significantly above chance level. We assess the linear classifier's performance using a bootstrap resampling scheme (k=100, stratified where relevant) and perform random permutations (1,000 permutations, 10 per bootstrap resampling) of the labels to obtain a null distribution of balanced accuracy and area under the receiver-operating curve (AUROC). We assess a CAV as significant if all metrics are higher than the estimated null distributions with $p<0.05$. We then estimate the generalizability of the classifier across time steps by performing the classification at all time points ($t=1,\dots,T$), where we give a label of 1 (resp. 0) at time points where the concept is present (resp. absent), when that information is known (i.e. synthetic data), and a label of 1 (resp. 0) for all time points of concept (resp. control) time series if not known (i.e. clinical application). This measure of performance beyond the $[t_{start}, t_{end}]$ window allows to understand whether concepts are represented similarly across all time points in the sequence, or whether the signal is specific to the window selected.

\paragraph{Presence of the concept in a sample:} The original TCAV work \citep{Kim2017} computes the cosine similarity between the activations $\vec{a}_l$ of a sample and the obtained CAV at each layer to estimate how similar an image is to a concept. This similarity measure can be thought of as estimating whether a concept is manifesting or ``present'' in the sample. In time series, it can be computed at each time point independently to obtain a (local) trajectory of concept presence per layer: 

\begin{align*}
\mathrm{tCA}_{C}(\vec{x}_t) = \frac{\vec{a}_{t}^T}{||\vec{a}_{t}||_2} \vec{v_C}
\end{align*} 

Where $\vec{v_C}$ corresponds to the unit norm CAV of concept $C$. This formulation can be extended to estimate whether the activations change over time in the direction of the concept by replacing $\vec{a}_{t}$ by $[\vec{a}_{t} - \vec{a}_{t-dt}]$ (following the assumption of local linearity in \citep{Kim2017}), where $dt$ represents a constant lag in a time shifting window. This is relevant to investigate concepts that would vary across time, e.g. by becoming more severe, and this the formulation used throughout this work.

\paragraph{Influence of the concept on the model's prediction:} ~\citet{Kim2017} define the Conceptual Sensitivity (CS), to estimate how the model's gradients align with the CAV. This quantity, when aggregated over samples, represents a global explanation. Mathematically, CS can be computed as the directional derivative:

\begin{align*}
    \mathrm{CS}_{C, l, t}(F, \vec{x}_t) & := \frac{\partial h(f_l(\vec{x}_t))}{\partial \vec{v_C}} \\
    & = \nabla h(f_l(\vec{x}_t))^T\vec{v_C}
    \label{eq:cs}
\end{align*}

Which amounts to computing the cosine similarity between the direction of the CAV and the model's gradients. In the present case, CS is computed at every time step of the local trajectory by taking the gradients of the models w.r.t. the sigmoid of the logits. The obtained scores can be aggregated over time and/or over samples to obtain global concept attributions. 

We believe that $tCA$ and CS can be seen as providing complementary information for global explanations, i.e. how is the presence/absence of the concept varying across time, and is the model influenced by the concept to make its decisions? Indeed, a concept being ``present'' does not guarantee that the model relies on it for prediction. On the other hand, a CS score of 0 means that affecting how present the concept is has (locally) no influence on the model output, but CS cannot reflect on whether the concept is present or absent.

\subsection{Synthetic timeseries}
\label{sec:methods_data}
Inspired by \citep{Goyal2019}, we evaluate the proposed approach on a synthetic dataset designed to isolate individual effects. 

\begin{figure}[!t]
\begin{center}
\centerline{\includegraphics[width=\columnwidth]{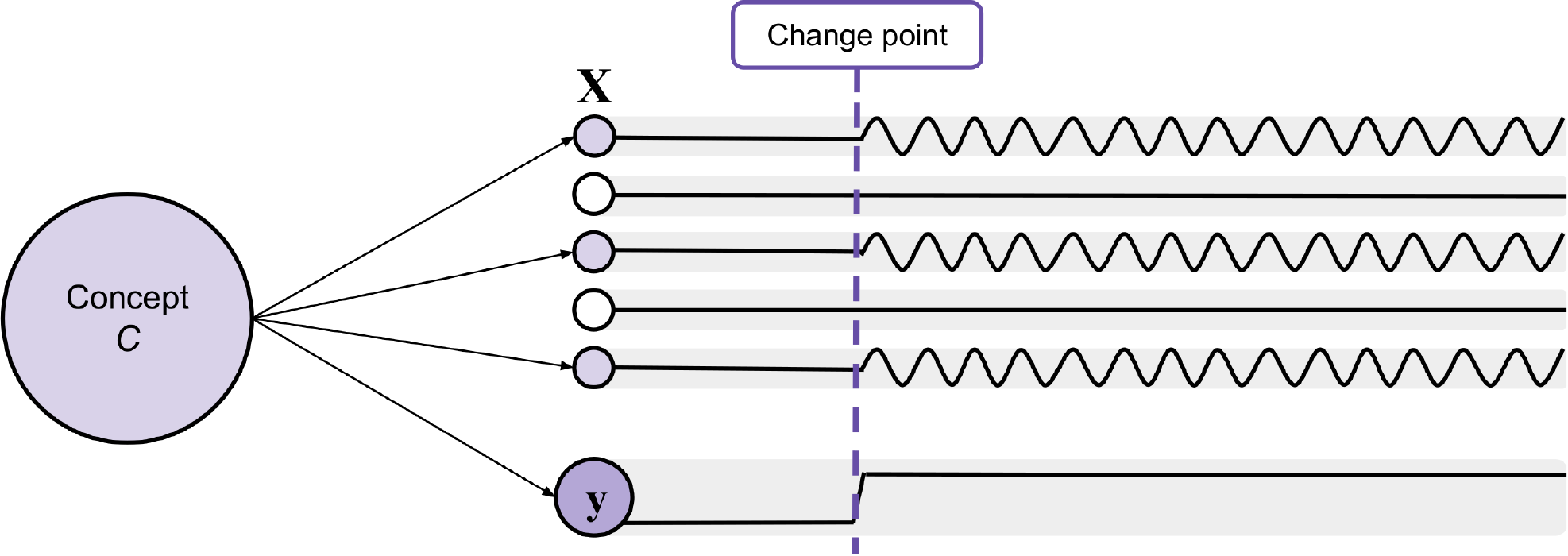}}
\caption{Illustration of the causal graph and sampling of time series for the synthetic dataset. A concept $C$ affects a subset of features $i \in {1,3,5}$ and a label $\vec{y}$ after the `change point'.}
\label{synthetic_data_causal_graph}
\end{center}
\end{figure}

\paragraph{Dataset design: }In our setup, a concept $C$, akin to a latent variable, can manifest through a causal relationship with a time series' features and label (see Figure \ref{synthetic_data_causal_graph}). For simplicity, we consider a binary behavior for $C$: a concept either manifests and is ``present'' in a sample after a selected temporal ``change point'', or it is ``absent''. When present, each feature has a predefined likelihood $p(d=1|C=1)=p(d=0|C=0)$ of exhibiting the concept's pattern, which can be any detectable change in behavior. If that likelihood is set to zero for a feature, the concept will not influence the feature's behavior. Similarly, the concept influences the label $y$ after the same change point with likelihood $p(y=1|C=1)=p(y=0|C=0)$. These parameters are set at the dataset level (see Supplement for the sampling algorithm).

In this work, we define two concepts, $C_1$ and $C_2$, and two corresponding labels, $y_1$ and $y_2$, influenced by $C_1$ and $C_2$ respectively with $p(y|C)=1$. We generate 10 numerical features with Gaussian background noise, and link $C_1$ and $C_2$ to non-overlapping sets of 5 features each. The pattern for all concept-activated features is the emergence of a sinusoid with fixed frequency and amplitude added to the Gaussian noise. The simplicity of this setup ensures that the ground truth is well understood. 

We note that our code supports more complex settings, e.g. overlapping concept-feature space or multi-concept label contingency tables (see Supplement). In addition, feature/concept behaviors can be made more realistic (e.g. by including binary variables). Therefore, this dataset is suitable for assessing attributions at both the feature and concept level, and could be used in other scenarios.

\paragraph{Model Training: }The model consists of a 3-layer stacked LSTM ~\citep{Hochreiter1997} RNN, with 64 hidden units contained in each layer. These layers are followed by a fully connected layer. The model is trained using cross entropy loss with the Adam optimizer (fixed learning rate of 3e-4) and batch size of 32, over 10000 randomly sampled batches. We report the model performance across all time steps and examples based on accuracy, AUROC and area under the precision-recall curve (AUPRC).

\paragraph{Concept definition: }We use $C_1$ and $C_2$ as our concepts. We assign $t_{start}$ as the change point and define the `time of interest' $t_{end}$ as 25 samples (arbitrary choice) after the change point, to ensure the concept is either present (concept group) or absent (control group). We randomly select 100 samples from the validation set to build a CAV for each concept (i.e. $C_1$ and $C_2$) and layer $l$. To ensure that the model is able to identify the concepts, we filter for a minimum model accuracy of 0.8 on a per-sequence basis (arbitrary threshold). The performance of each CAV is assessed on the held-out time steps during the bootstrap procedure, as well as on 500 other time series of the validation set (all time steps).

\subsection{Illustration on clinical predictions}
\label{sec:methods_mimic}
\paragraph{Data: }We use the de-identified critical care EHR data from the Medical Information Mart for Intensive Care (MIMIC-III) ~\citep{Johnson2016a, Johnson2016, Goldberger2000} to investigate a real-world application of our technique. After filtering out patients under the age of 18, the MIMIC-III dataset contained 47,296 patients, which were randomised  across training (80\%), validation (10\%), and test (10\%) sets. Each patient's medical history is converted to a time series of one-hour aggregates including different structured data elements (medication, labs, vitals, \dots) represented by numerical and binary variables \citep[see Supplement for details]{Tomasev2019}. Importantly, our data representation is sparse at each time step, and includes 32,170 continuous variables and 38,600 binary variables, for a total of 70,770 features.

\paragraph{Model: }We focus on the predictions 48 hours in advance of an Acute Kidney Injury (AKI) event of stage 1 or more ~\citep[max stage 3, as per the Kidney Disease Improving Global Outcomes classification, KDIGO, ][]{Khwaja2012}. We use the same model architecture as described in ~\citep{Tomasev2019} which consists in a 3-layer stacked RNN with residual connections and add dropout probability of 0.4 to the output connections of each LSTM cell. The model's hyper-parameters were defined based on a grid search on the validation set. The model's performance is then assessed on the test set using AUPRC given the low prevalence of AKI in the dataset. For comparison with the literature, we also report AUROC.

\paragraph{Concept definition: }Based on clinical input, we define illustrative concepts by relying on rule-filtering of specific clinical events from patients included in the validation set. These events then serve to determine $t_{start}$ and $t_{end}$ for the different CAV building strategies. We define multiple concepts:
\begin{itemize}
    \item `AKI', that is directly related to the outcome labels for sanity check.
    \item `Nephrotoxicity', a known risk factor for kidney injury.
    \item `Antibiotics'. This concept aims at identifying bacterial infections, a proxy for sepsis, which is another known risk factor for kidney injury.
    \item `Sex'.
\end{itemize}

The AKI concept group is defined as follows: admissions where the patient is recorded to have normal renal function \citep[i.e. no AKI, based on the serum creatinine and the KDIGO criteria][]{Khwaja2012}, and later in the admission renal function degrades to an AKI stage 2 ($t_{end}$). The control group for the AKI concept is defined as: admissions where no AKI is recorded, with at least one normal creatinine measurement. In this case, a random one-hour bucket is selected as representing the end point of the trajectory, i.e. $t_{end}$. The AKI concept is purposefully circular to assess how the proposed approach scales to real-world problems.

To define the nephrotoxicity concept, we select admissions where the patient has normal renal function, then receives a particular type of nephrotoxic agent, which is followed by an AKI stage 1, 2 or 3. Please note that we do not select admissions where the nephrotoxic agent \textit{caused} the AKI, as we do not have that information. We selected one class of nephrotoxic drugs, non-steroidal anti-inflammatory drugs (NSAIDs). The control group for this concept included admissions where the patient had normal renal function followed by an AKI (stage 1, 2 or 3), without receiving an NSAID medication before the adverse event\footnote{Please note that we control for the endpoint of the trajectory $t_{end}$ to correspond to an AKI event. One could however control for the proportion of AKI samples across both groups. On the other hand, not controlling for the endpoint might introduce a confounding factor if patients receiving NSAIDs have a higher prevalence of AKI.}. This `relative' concept \citep{Kim2017} investigates whether the model encodes the \textit{differential} effect of NSAIDs on AKI, compared to all other factors correlating with or causing AKI.

We identified 18 antimicrobial agents (see Supplement) and defined an `antibiotics' concept to act as a proxy for detecting sepsis. The selection of admissions for CAV building is similar to that of the nephrotoxicity concept.

The `sex' concept investigates whether self-reported sex affects the model's predictions, as the authors of \citep{Tomasev2019} reported lower performance of the model on women compared to men. To this end, we build a CAV distinguishing between admissions of females and males leading to an AKI episode. We consider time windows spanning 12 hours before the AKI event, 24 hours before the AKI event or the beginning of the admission to the AKI event. Importantly, sex is not included as a feature in the model training.

To avoid potential confounding factors in the CAV, we use the same number of patients in the concept and control groups and choose patients to match on selected data statistics between groups. The features we match on are age, gender, duration of hospital admission, time between admission and AKI (or time between admission and selected $t_{end}$ for controls without AKI events), and inpatient mortality. For each of the patients in the concept and control groups we calculate the vector of features that we wish to match on, standardising based on the training dataset mean and standard deviation.  From the pool of candidate examples for the control group, we then select those that minimise the total L1 distance between feature vectors in the control and concept group. The distance minimisation problem is solved using the Munkres algorithm \citep{Munkres1957}. We then select similar numbers of time steps within each patient (arbitrarily selected as 10), based on different sampling strategies: random sampling across the $t_{start}:t_{end}$ window, sampling at equal intervals with the interval being computed based on the number of time steps for a patient $T_i$, and sampling among the true positive predictions. We report results for sampling across equal intervals, and did not identify a significant effect of the sampling strategy on the results.

We build CAVs for each concept based on selected concept and control groups from the training set. We then compute $tCA$ and CS on patients from similarly selected groups from the test set.

\paragraph{Comparison with feature-based attributions:} We present the results of occlusion \citep{Zeiler2014} and gradient \citep{Simonyan2013} analyses, computed for features that are `present' in a time step. These attribution scores are estimated at each time step independently, as in \citep[see Supplement for details]{Tomasev2019}. The aim of these analyses is to highlight the differences between feature-based and concept-based techniques. We however believe that both can potentially be useful, and do not intend to recommend one over the other.

\section{Results}
\subsection{Synthetic dataset}

\begin{figure*}[!ht]
\centering
\begin{subfigure}[t]{0.03\textwidth}
\textbf{a}
\end{subfigure}
\begin{subfigure}[t]{0.35\textwidth}
\includegraphics[width=0.9\linewidth,valign=t]{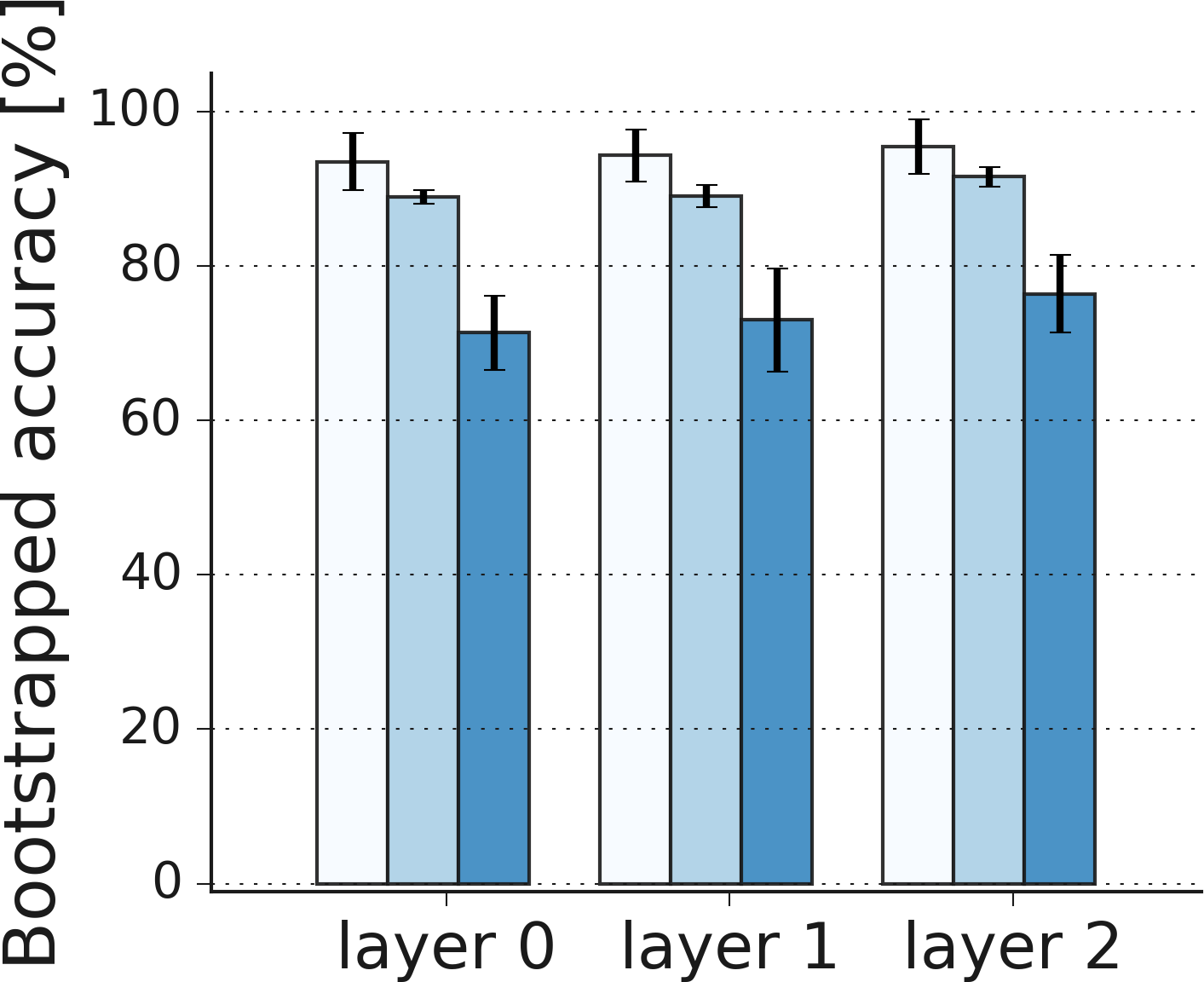} 
\end{subfigure}
\begin{subfigure}[t]{0.03\textwidth}
\textbf{b}
\end{subfigure}
\begin{subfigure}[t]{0.35\textwidth}
\includegraphics[width=0.9\linewidth,valign=t]{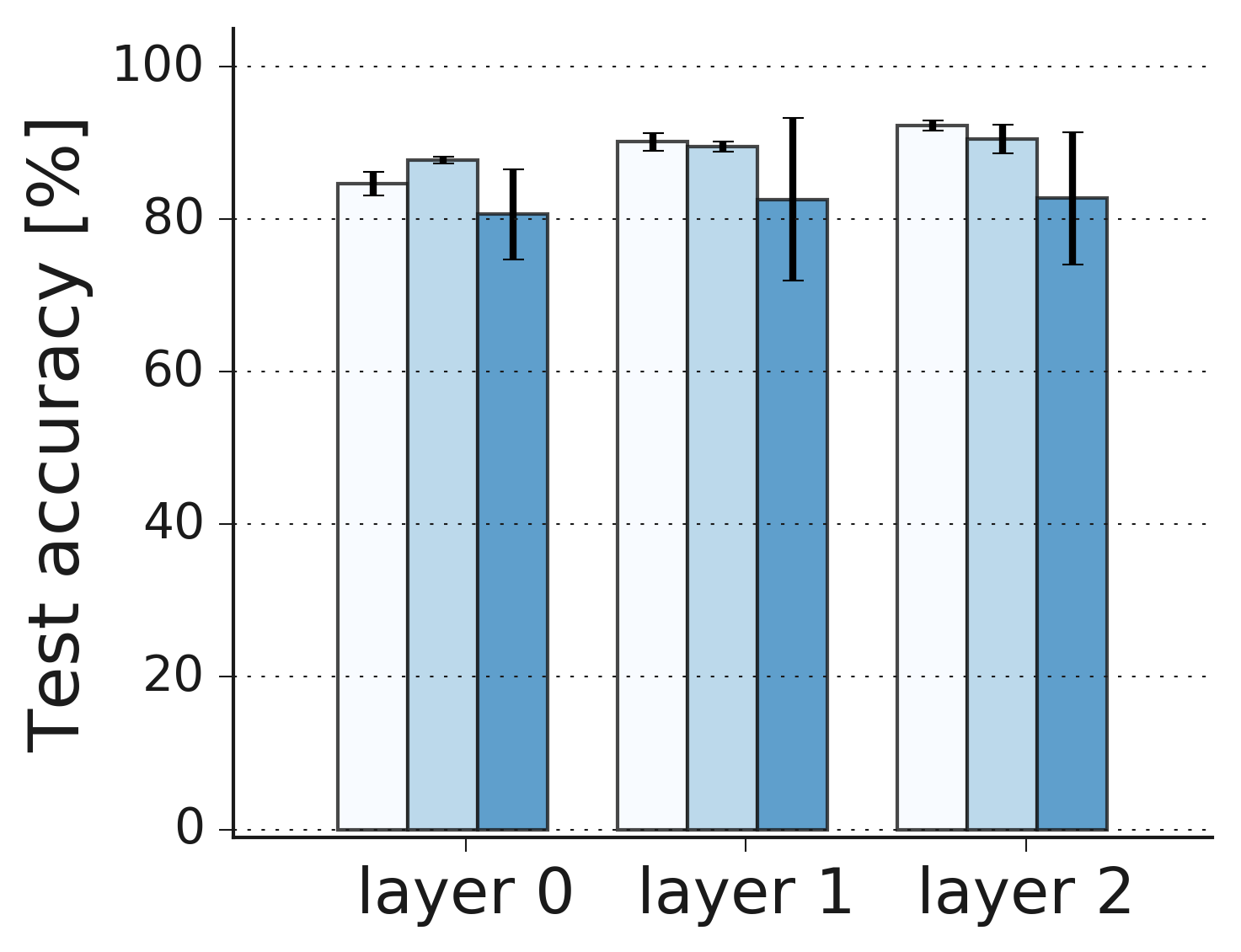}
\end{subfigure}
\begin{subfigure}[t]{0.2\textwidth}
\includegraphics[width=0.5\linewidth, valign=t]{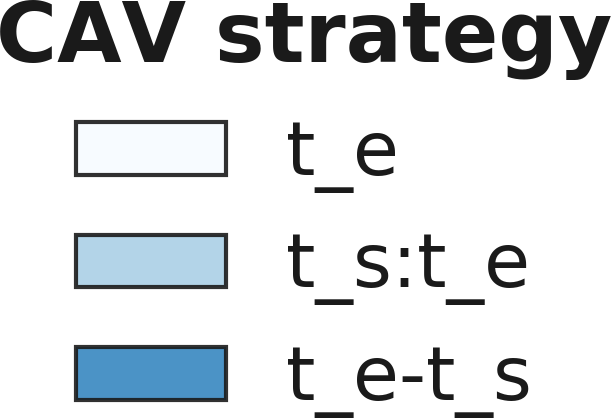}
\end{subfigure} \\
\begin{subfigure}[t]{0.03\textwidth}
\textbf{c}
\end{subfigure}
\begin{subfigure}[t]{0.3\textwidth}
\includegraphics[height=4cm, valign=t]{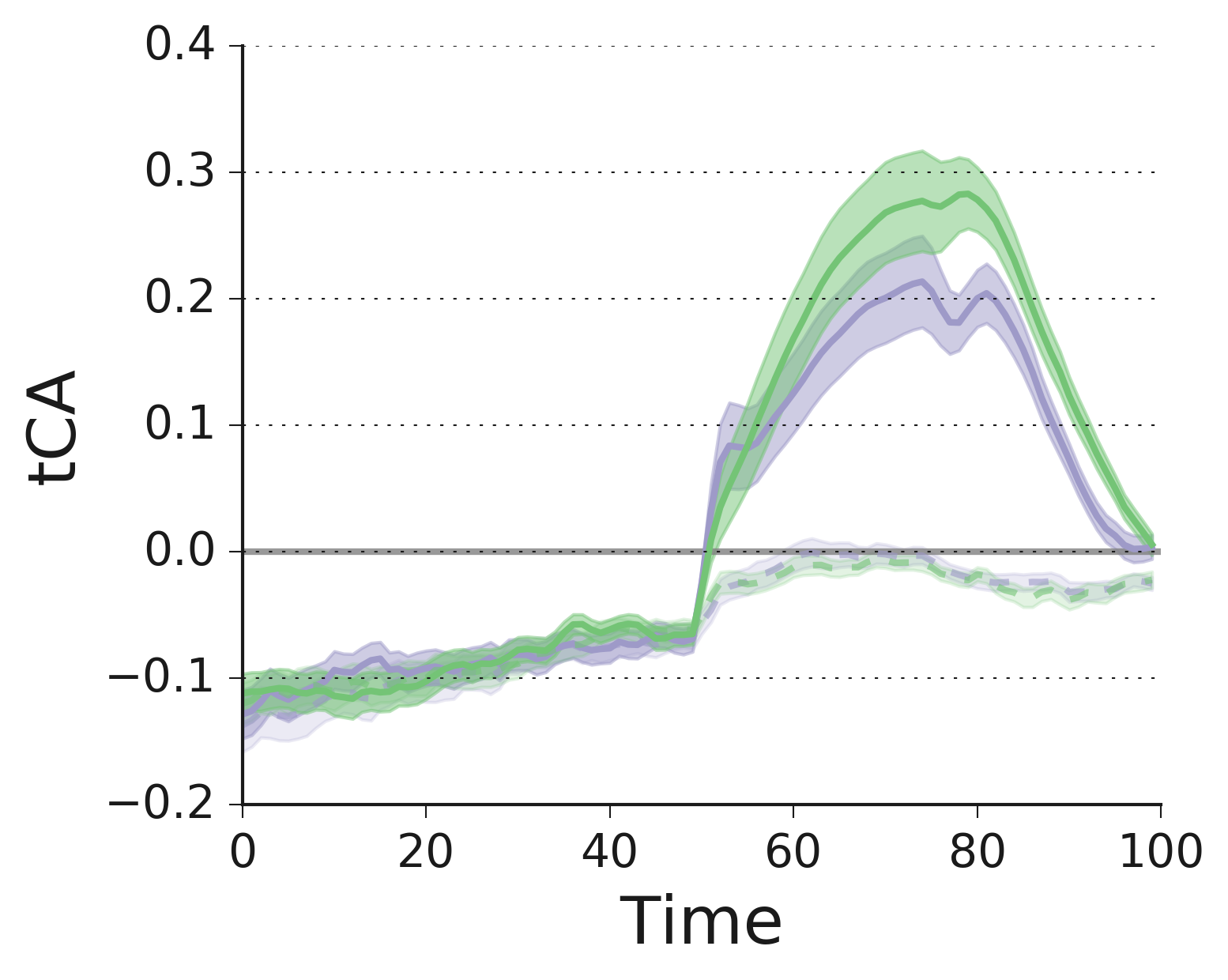}
\end{subfigure}
\begin{subfigure}[t]{0.03\textwidth}
\textbf{d}
\end{subfigure}
\begin{subfigure}[t]{0.6\textwidth}
\includegraphics[height=4cm, valign=t]{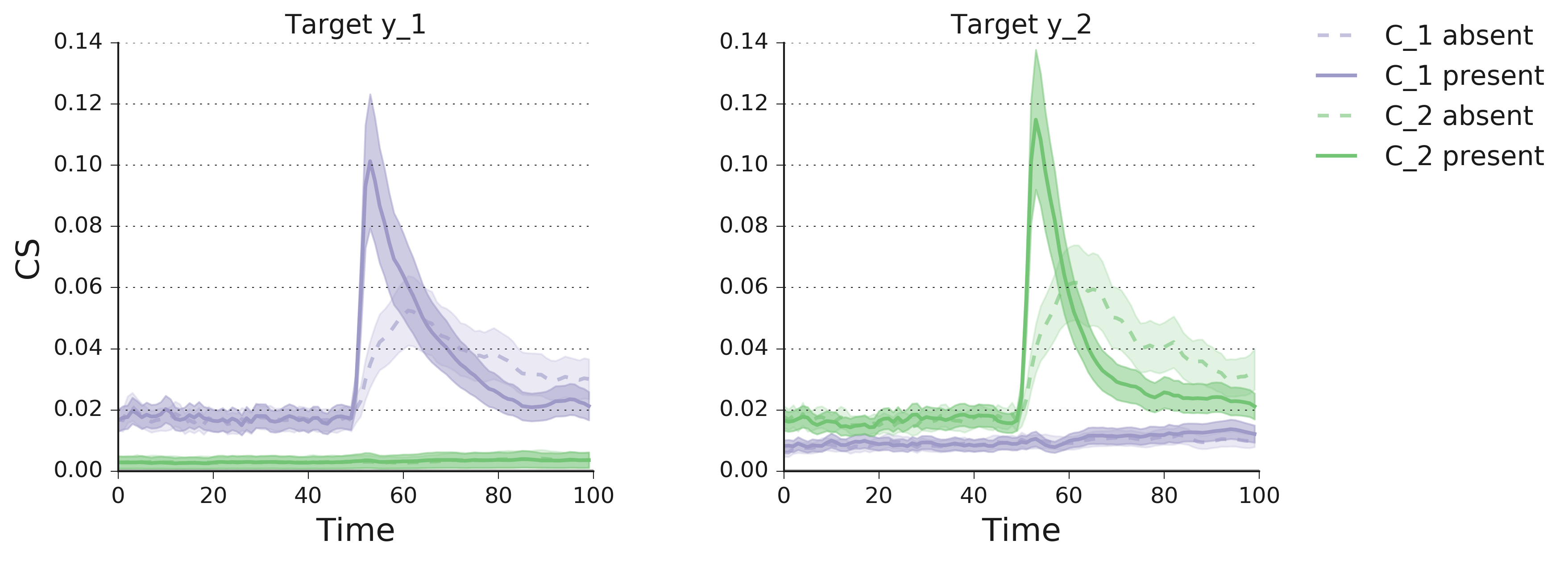}
\end{subfigure}
\caption{\textbf{Synthetic data results.} Accuracy of the $C_1$ CAV in bootstrap (a) and test (b) evaluations for CAV$_{t_{end}}$, CAV$_{t_{start}:t_{end}}$ and CAV$_{t_{end}-t_{start}}$, in \%. c $tCA$ scores (layer 2) averaged across time series, and bootstraps (mean$\pm$std) when a concept ($C_1$ green, $C_2$ purple) is absent (light shade, dotted line) or present (dark shade). All timeseries are aligned to have their changepoint at $t=50$. d CS scores for target $y_1$ (left) and $y_2$ (right).}
\label{fig:synth_cavs}
\end{figure*}

% \begin{figure*}[tb]
% \centering
% \begin{subfigure}[t]{0.03\textwidth}
% \textbf{a}
% \end{subfigure}
% \begin{subfigure}[t]{0.3\textwidth}
% \includegraphics[height=4cm, valign=t]{figures/final/tCA_global_timeseries.png}
% \end{subfigure}
% \begin{subfigure}[t]{0.03\columnwidth}
% \textbf{b}
% \end{subfigure}
% \begin{subfigure}[t]{0.63\textwidth}
% \includegraphics[height=4cm, valign=t]{figures/final/CS_global_timeseries.png}
% \end{subfigure}
% \caption{\textbf{Concept influence. a} $tCA$ scores (layer 2) averaged across time series, and bootstraps (mean$\pm$std) when a concept ($C_1$ green, $C_2$ purple) is absent (light shade, dotted line) or present (dark shade). All timeseries are aligned to have their changepoint at $t=50$. \textbf{b} CS scores for target $y_1$ (left) and $y_2$ (right).}
% \label{fig:synth_main}
% \end{figure*}

\paragraph{Data and model:} We generate 10,000 time series of 100 time points each to predict $y_1$ and $y_2$. After training, the model reaches 95.34\% accuracy, 0.8511 AUPRC and 0.9274 AUROC on a test set of 1,000 time series.

\subsubsection{Building the CAV} 
The different strategies lead to significant CAVs for both $C_1$ and $C_2$ as assessed on held-out test sets, although CAV$_{t_{end}-t_{start}}$ has relatively lower performance (see Figure \ref{fig:synth_cavs}a for $C_1$ and Supplement). All CAVs generalize to time points outside of the $[t_{start}:t_{end}]$ time window used for building, on further validation time series (Figure \ref{fig:synth_cavs}b). We however note that this result might be driven by the simplicity of our synthetic dataset and the high performance of the RNN model. For compactness, further results focus on the CAV$_{t_{start}:t_{end}}$ strategy. All strategies however lead to similar results in terms of CS and $tCA$ scores.

\subsubsection{Presence of the concept over time}
$tCA$ is estimated at each time step, using a lag of 25 time steps (arbitrary choice) for both concepts. Figure \ref{fig:synth_cavs}c displays the average across aligned time series. We observe that $tCA$ has a negative score when the concept is absent, and then sharply transitions to positive scores when the concept becomes present ($t=50$).

\subsubsection{Influence of the CAV}
We compute CS at each time point and display global trajectories of the obtained scores in Figure \ref{fig:synth_cavs}d. The results display the expected behavior: for target $y_2$, only concept $C_2$ has CS scores that are not tightly distributed around zero at all time points. In addition, CS scores are low before the change point (here aligned across all time series as $t=50$), reflecting the ``absence'' of $C_1$, while they become positive at the change point, when the label and concept manifest. These results are replicated for target $y_1$ and $C_1$ (see Supplement).

\subsection{MIMIC dataset}
\subsubsection{Data and model} Our model predicts AKI of any severity within the next 48 hours with a AUPRC of 0.491 and AUROC of 0.798. It is difficult to make direct comparisons with the literature as, to our knowledge, comparable continuous AKI predictions on MIMIC have not been reported on to date. However, there are a number of similar studies on different EHR datasets: ~\citet{simonov2019} report an AUROC of 0.74 for AKI within 24 hours using a discrete time logistic regression triggered after every new measurement; while ~\citet{Kate2020} report up to 0.724 with a similar setup. ~\citet{Flechet2019} predict AKI within the next 7 days in an ICU population with AUROC ranging from 0.80-0.95 depending on the window of input data.

\subsubsection{AKI concept}
\begin{figure*}[!ht]
\begin{center}
\includegraphics[width=0.85\textwidth]{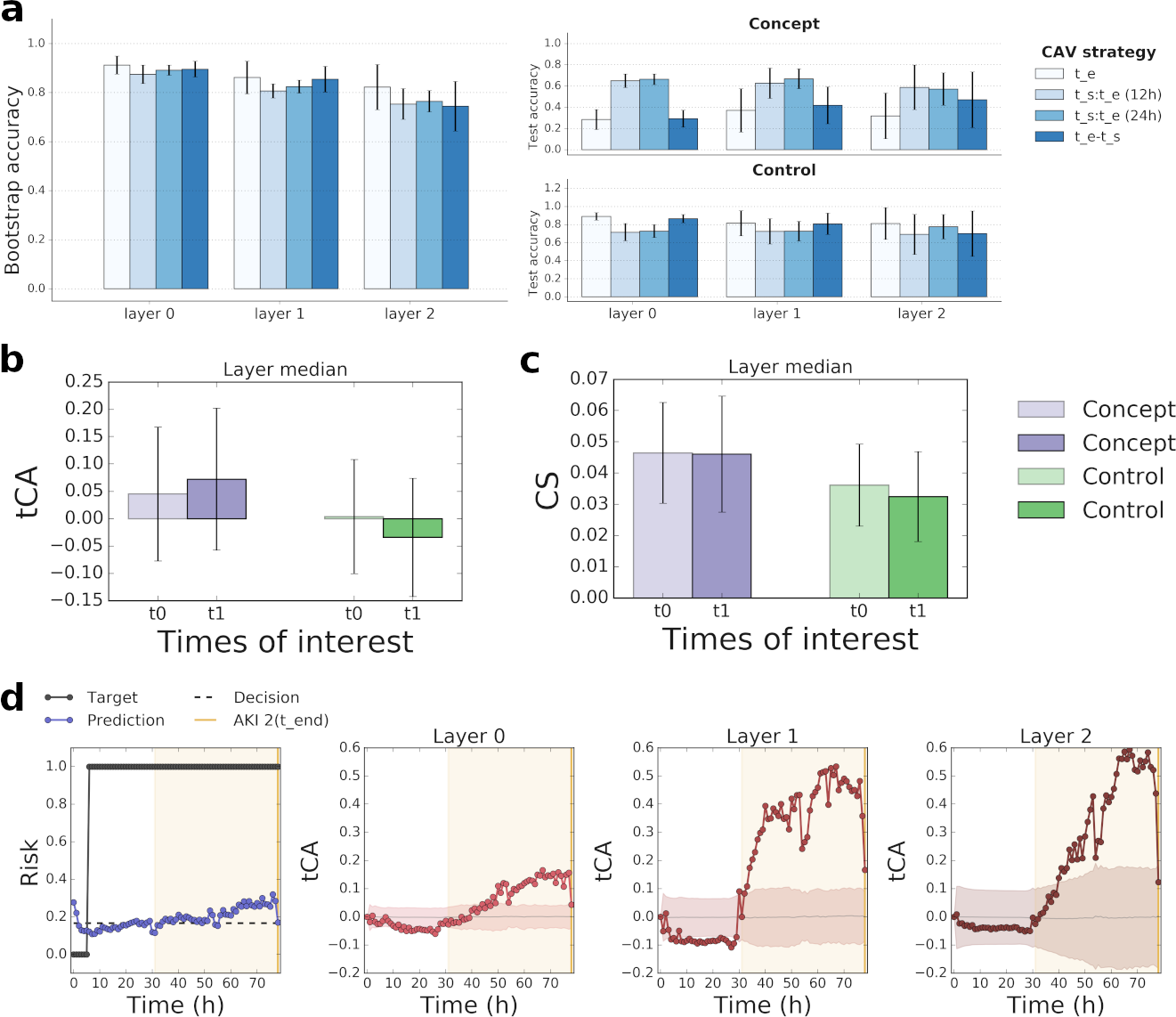}
\caption{\textbf{MIMIC results, AKI concept. a } CAV linear model performance on held-out samples (left) and on test samples for concept (top) and control (bottom) samples. Each bar represents a CAV building strategy. \textbf{b} $tCA$ global scores, averaged across patients (mean$\pm$std) for the concept (purple) and control (green) groups 48 hours before AKI ($t_0$) and at time of AKI stage 2 ($t_1$). \textbf{c} $CS$ scores. \textbf{d} Single patient timeseries, displaying the label (AKI 1+ within 48h) and model's output, as well as $tCA$ for each layer, and its null hypothesis (shaded). The yellow shaded area represents the prediction horizon of the model, i.e. within 48 hours of AKI 2.}
\label{fig:mimic_aki}
\end{center}
\end{figure*}

\paragraph{Building the CAV:} The AKI concept was built using 161 in-patient episodes selected from the validation set. We tested different building strategies, namely CAV$_{t_{end}}$ with $t_{end}$ being the time of AKI, CAV$_{t_{start}:t_{end}}$ using time steps included in the 12 or 24 hours before AKI and CAV$_{t_{end}-t_{start}}$ by subtracting the activation at time of admission from that of the time of AKI. The obtained linear classifier was then evaluated on examples selected from the test set, on the same points as used for training (i.e. $t_{end}$, last 12 hours, ...) as well as on all other time points. We observe high training accuracy for all models, as evaluated per the bootstrap scheme (Figure \ref{fig:mimic_aki}a (left), Supplement). When evaluating on the test examples, we observe high accuracy on the equivalent time steps as used for training. This accuracy decreases for the concept group when evaluating on test time series (all time steps), with CAV$_{t_{end}}$ and CAV$_{t_{end}-t_{start}}$ seemingly overfitting to the training time steps (Figure \ref{fig:mimic_aki}a (right)). This result suggests that time steps outside of the selected window might not reflect the same signals as time steps within the selected window.

\paragraph{Presence of the concept over time:} We select the model with best generalization across time steps to compute the alignment, i.e. using the last 24 hours before AKI and compute $tCA$ with a lag $dt$ of 24 hours. We present the results using the test examples at time 48 hours before $t_{end}$ (our prediction horizon, $t_0$ on Figure \ref{fig:mimic_aki}b (left)) and at time $t_{end}$ (corresponding to time of AKI stage 2 for the concept group and a random no AKI event for the control group, $t_1$).

As previously observed on the synthetic data, using the difference in activations on a 24 hour sliding window leads to higher $tCA$ scores when the concept is present compared to when it is absent, with an increase from $t_{start}$ to $t_{end}$. We plot the time series of concept alignment for an example patient in Figure \ref{fig:mimic_aki}d for each layer of the model, along with the distribution of $tCA$ scores when using permuted CAV vectors (i.e. built from randomized concept/control labels) for an example from the evaluation group with AKI. The $tCA$ scores per time step seem to reflect the risk as predicted by the model: the $tCA$ score starts increasing around time $t~30$, i.e. around 48 hours before the AKI stage 2 event detected by the KDIGO label.

\paragraph{Influence of the CAV:} We observe a positive influence of the CAV on the target, as displayed by strictly positive values of CS on the different samples and time points considered (Figure \ref{fig:mimic_aki}c). This is expected due to the circularity in the definition of this concept.

\subsubsection{NSAIDs concept}
\begin{figure*}[!ht]
\begin{center}
\includegraphics[width=0.85\textwidth]{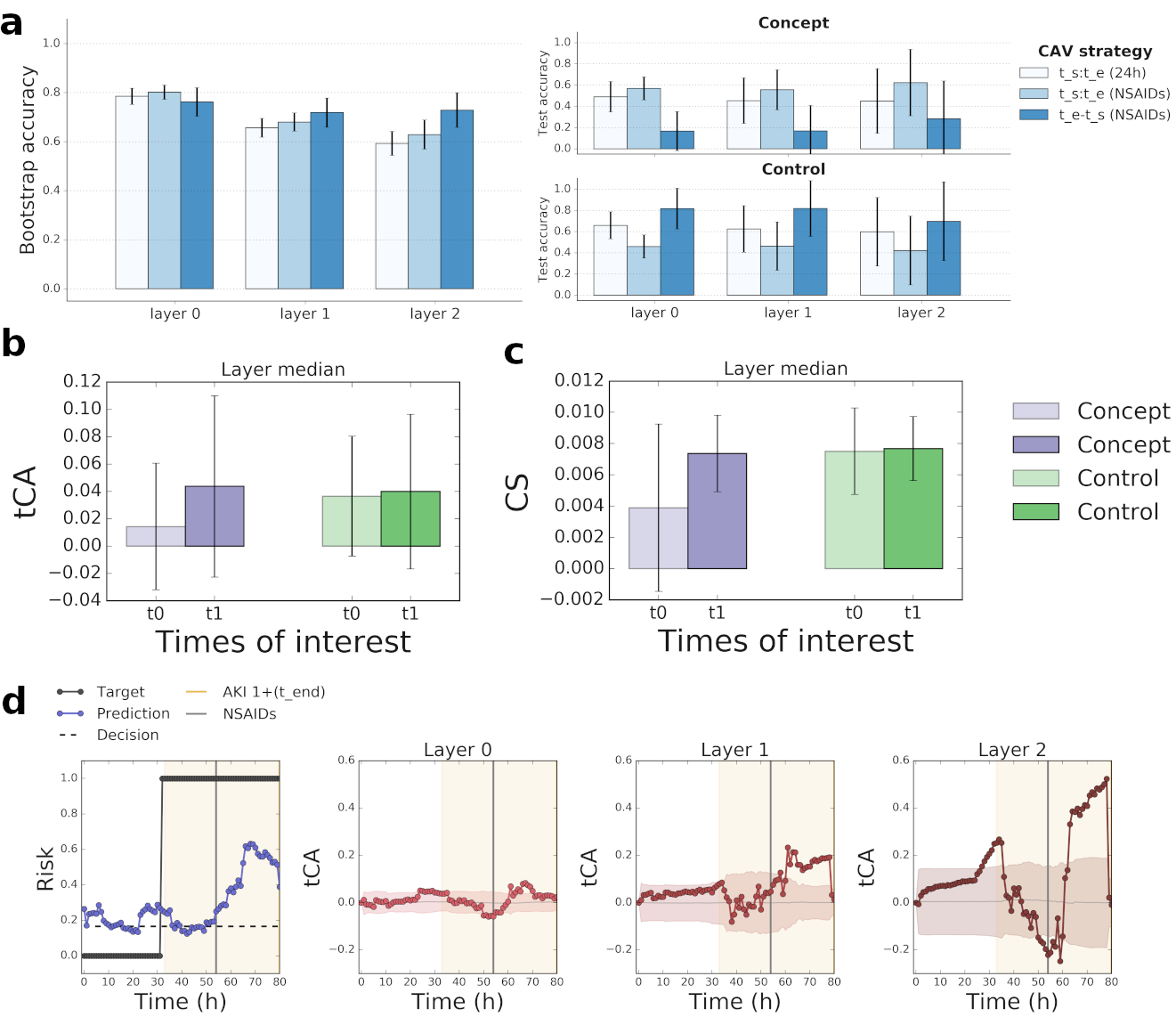}
\caption{\textbf{MIMIC results, NSAIDs concept. a } CAV linear model performance on held-out samples (left) and on test samples for concept (top) and control (bottom) samples. Each bar represents a CAV building strategy. \textbf{b} $tCA$ global scores, averaged across patients (mean$\pm$std) for the concept (purple) and control (green) groups 48 hours before AKI ($t_0$) and at time of AKI stage 2 ($t_1$). \textbf{c} $CS$ scores. \textbf{d} Single patient timeseries, displaying the label (AKI 1+ within 48h) and model's output, as well as $tCA$ for each layer, and its null hypothesis (shaded). The yellow shaded area represents the prediction horizon of the model, i.e. within 48 hours of AKI 1+. The administration of NSAIDs is displayed by a grey vertical line.}
\label{fig:mimic_nsaids}
\end{center}
\end{figure*}

\paragraph{Building the CAV:} Similarly to the AKI concept, we train three variants of the linear classifier\footnote{Given that the end point (AKI 1+) is controlled for, CAV$_{t_{end}}$ is not built.}, where we use all samples between 24 hours before AKI to the time of AKI ($t_{end}$), all samples between the time of NSAIDs and the time of AKI, or the difference in activation between the time of AKI and the time of NSAIDs. While all classifiers are assessed as significant per non-parametric permutation testing, model performance is overall lower than for the AKI concept both for the held-out and validation time steps (Figure \ref{fig:mimic_nsaids}a and Supplement). As previously, we select CAV$_{t_{start}:t_{end}}$ with $t_{start}$ being 24 hours before AKI to evaluate $tCA$ and $CS$ scores.

\begin{figure*}[tb]
\centering
\begin{subfigure}[t]{0.05\textwidth}
\textbf{a}
\end{subfigure}
\begin{subfigure}[t]{0.8\textwidth}
\includegraphics[width=\linewidth, valign=t]{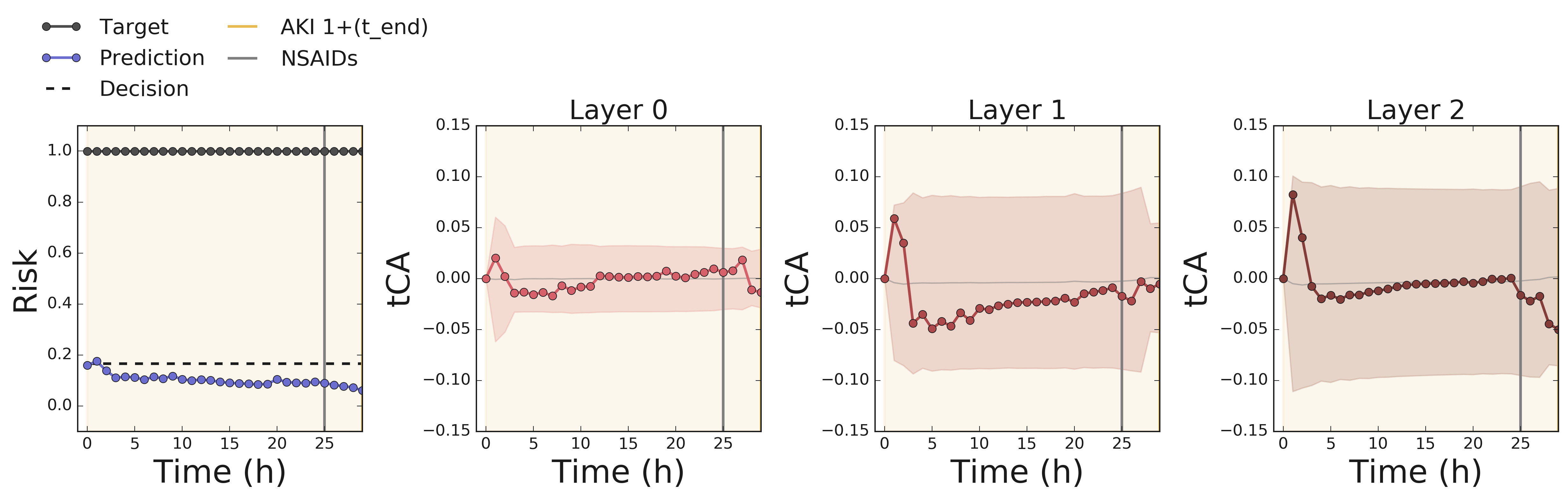}
\end{subfigure}  \\
\begin{subfigure}[t]{0.05\textwidth}
\textbf{b}
\end{subfigure}
\begin{subfigure}[t]{0.8\textwidth}
\includegraphics[width=\linewidth, valign=t]{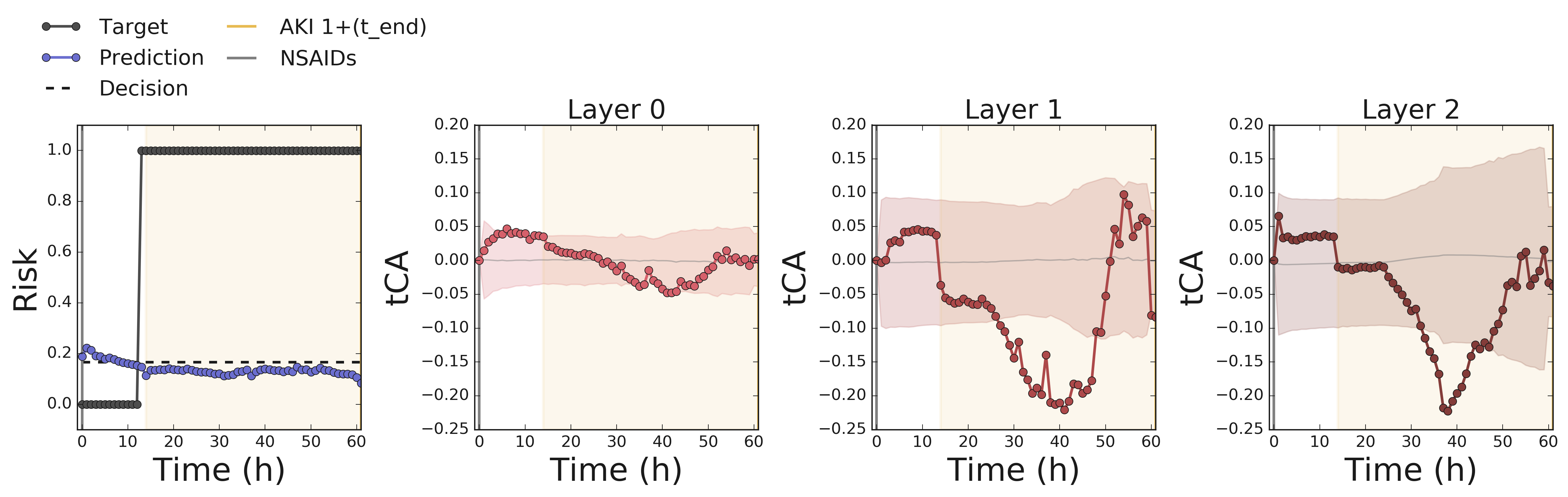}
\end{subfigure} \\
\begin{subfigure}[t]{0.05\textwidth}
\textbf{c}
\end{subfigure}
\begin{subfigure}[t]{0.8\textwidth}
\includegraphics[width=\linewidth, valign=t]{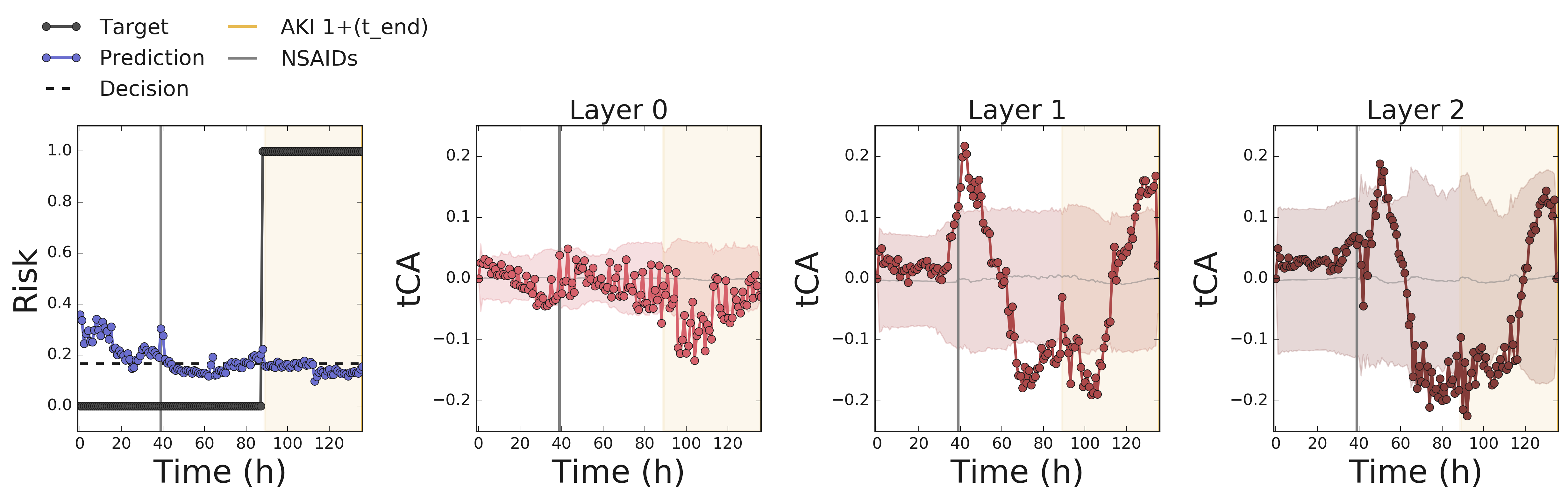}
\end{subfigure}
\caption{Local MIMIC results, NSAIDs concept. a,b Negative predictions, c false positive prediction at time of NSAIDs. Similar to Figure~\ref{fig:mimic_nsaids}d.}
\label{fig:mimic_nsaids_local_TN}
\end{figure*}

\paragraph{Presence of the concept over time:} When using a 24-hour sliding window of activation differences, we obtain higher $tCA$ scores directly before the AKI event ($t_1$, 2 hours before AKI) when the concept is present compared to at time of NSAIDs ($t_0$, Figure \ref{fig:mimic_nsaids}b). This difference is however small and similar scores are obtained close to the AKI endpoint on control patients. Figure~\ref{fig:mimic_nsaids}d displays an example trajectory from the evaluation set for CAV$_{t_{start}:t_{end}}$. We discern an increase in alignment, outside of the $\pm 1 \times$ standard deviation, after the time of NSAIDs administration on the three layers. See the Supplement for more positive local examples. We observe no or negative alignment for negative predictions (Figure~\ref{fig:mimic_nsaids_local_TN}, a,b). On the other hand, we observe an increase in alignment at time of NSAIDs administration, simultaneous to a false positive prediction (around $t=40$, Figure~\ref{fig:mimic_nsaids_local_TN}c).

\paragraph{Influence of the CAV:} Consistent with our observations, CS displays a small effect of the concept on the predictions at the time of AKI, but the pattern is not as clear as for the AKI concept (Figure \ref{fig:mimic_nsaids}c). This could reflect either that the CAV does not properly represent the NSAIDs direction, or that the model is not only marginally influenced by this direction when making predictions. Further work will investigate other nephrotoxic agents as well as involve a clinical evaluation of patients to ensure that the agent \textit{caused} the AKI.

\subsubsection{Antibiotics concept}
\paragraph{Building the CAV:} Given previous results, we select CAV$_{t_{start}:t_{end}}$ with $t_{start}$ being 24 hours before AKI to build CAVs and evaluate $tCA$ and $CS$ scores. The classifiers are assessed as significant per non-parametric permutation testing, with accuracies of 61.04\%, 64.77\% and 63.38\% and ROC AUC of 0.6891, 0.7275 and 0.7325 for layers 1, 2 and 3, respectively ($p<0.001$).

\begin{figure*}[tb]
\begin{subfigure}[t]{0.05\textwidth}
\textbf{a}
\end{subfigure}
\begin{subfigure}[t]{0.4\textwidth}
\includegraphics[width=\linewidth,valign=t]{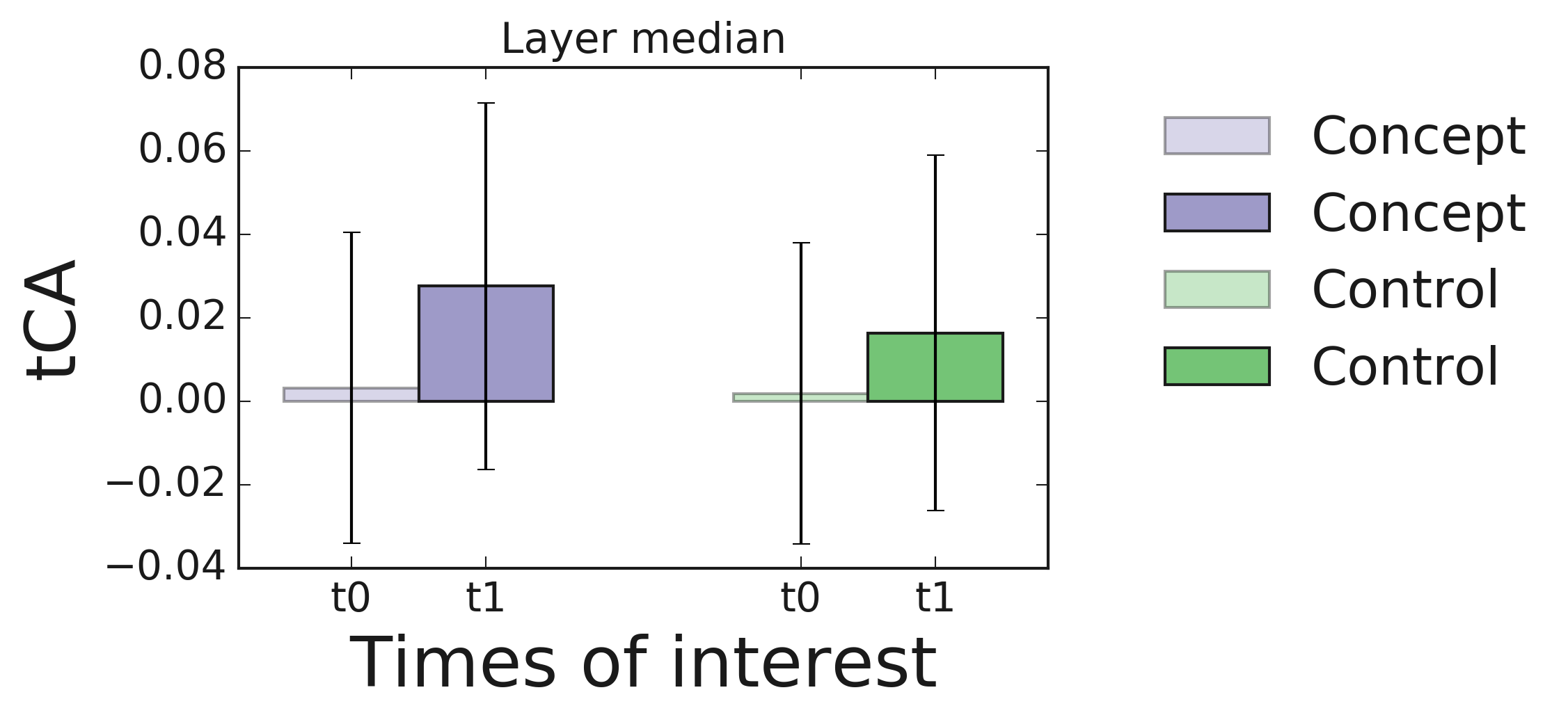}
\end{subfigure}
\begin{subfigure}[t]{0.05\textwidth}
\textbf{b}
\end{subfigure}
\begin{subfigure}[t]{0.4\textwidth}
\includegraphics[width=\linewidth,valign=t]{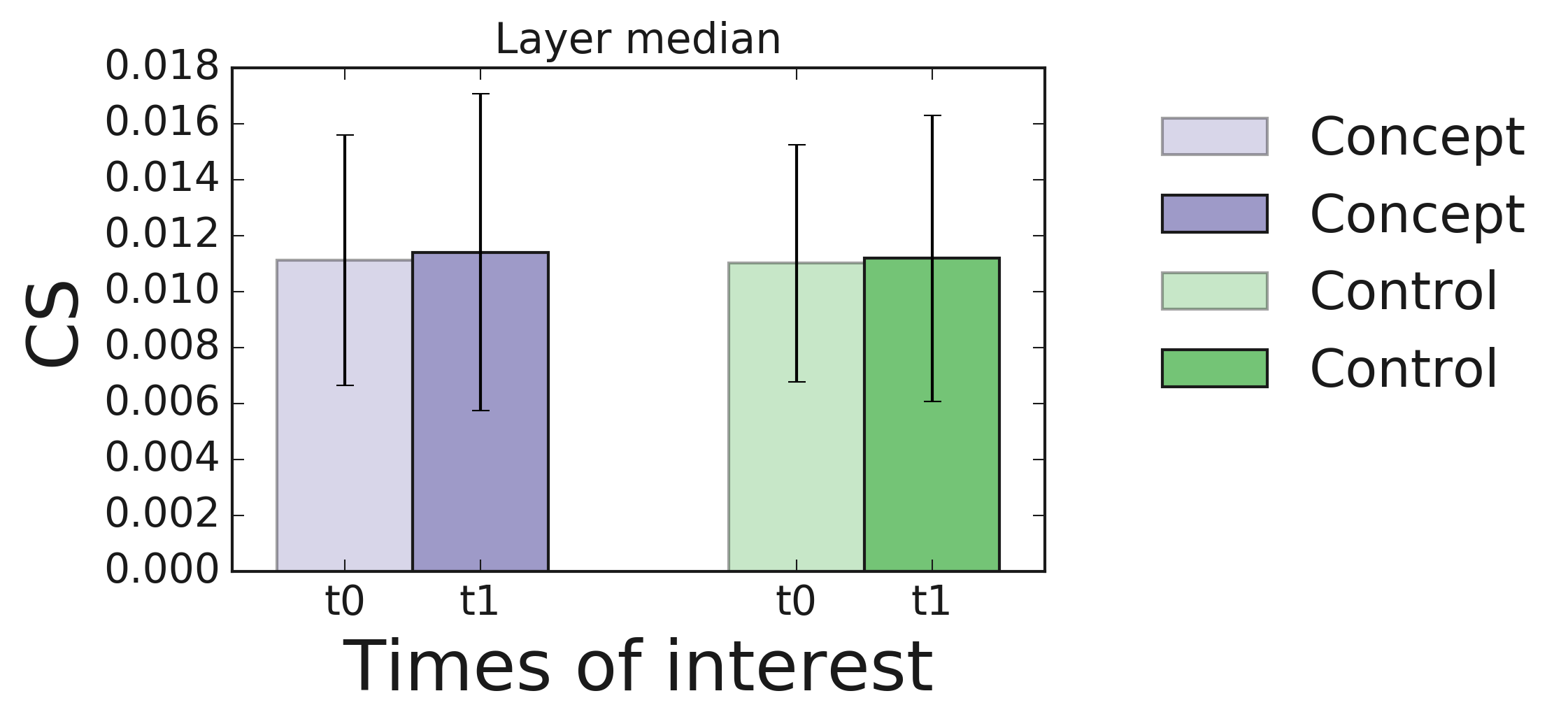}
\end{subfigure} \\
\begin{subfigure}[t]{0.05\textwidth}
\textbf{c}
\end{subfigure}
\begin{subfigure}[t]{0.85\textwidth}
\includegraphics[width=0.9\linewidth, valign=t]{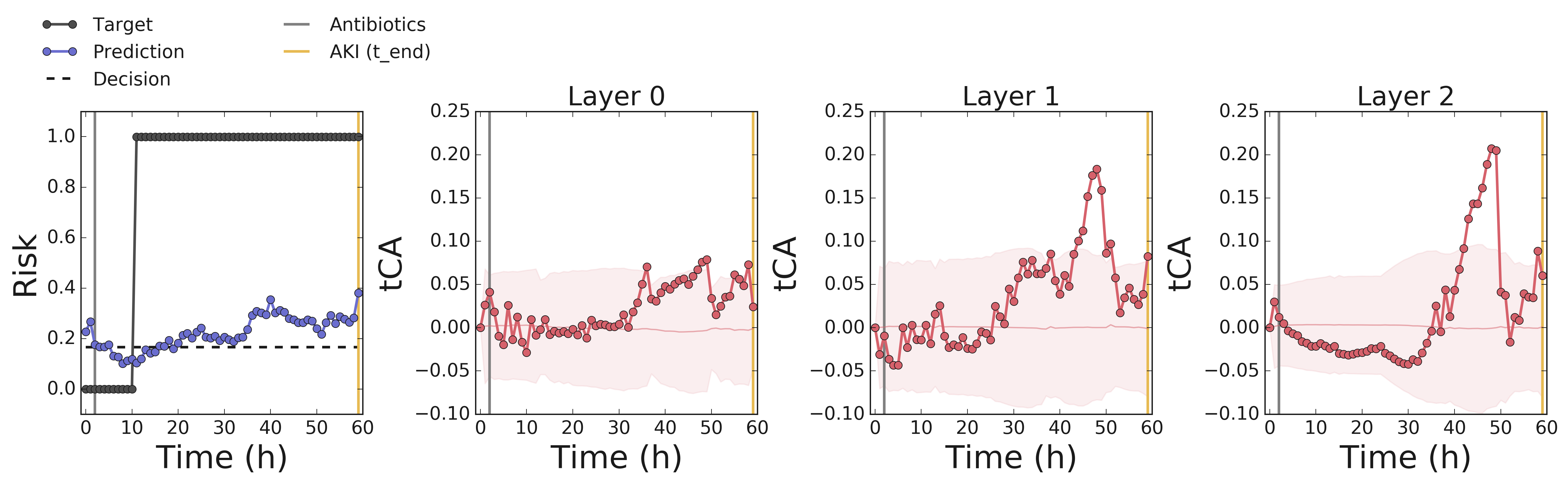}
\end{subfigure}
\caption{\textbf{MIMIC results, Antibiotics concept.} tCA (a) and CS (b) results at time of antibiotics, and at time of AKI. c Local patient trajectory with time of antibiotics displayed as a vertical grey line.}
\label{fig:mimic_antibio_global}
\end{figure*}

\paragraph{Presence of the concept over time:} When using a 24-hour sliding window of activation differences, we obtain higher $tCA$ scores at $t_1$ when the concept is present compared to at time $t_{start}$ (Figure~\ref{fig:mimic_antibio_global}). As for the NSAIDs concept, we observe increases in local tCA scores when the concept is present (Figure~\ref{fig:mimic_antibio_global} and Supplementary).

\paragraph{Influence of the CAV:} As per CS, the `antibiotics' concept has an influence on AKI predictions (Figure~\ref{fig:mimic_antibio_global}b).

\subsubsection{Sex concept}
None of the considered strategies to build CAVs leads to significant results, with balanced accuracy ranging between 0.4727 and 0.5905 ($p=0.19$), and ROC AUC between 0.4516 and 0.6190 ($p=0.18$, see Supplement for details). Interestingly, model performance on subgroups display no striking imbalance between male and female subgroups (0.5953 PRAUC for females and 0.5809 for males, with ROCAUC of 0.8624 and 0.8445 respectively). This result could suggest that features related to AKI are not significantly different between sexes, or that the model does not encode sex to predict AKI using the MIMIC benchmark dataset. On the other hand, non-significant CAVs could also arise from technical factors such as, e.g. the patient selection being too heterogeneous, or the time window selected for CAV building not including the feature changes between sexes. Therefore, a non-significant CAV does not allow to conclude that the model is not relying on the considered signal.

\subsubsection{Comparison with feature-based attributions}
Table~\ref{tab:mimic_aki_occlusion_gradients} displays the top 15 features ranked by their attribution scores, computed from the test set. We observe little overlap between the two lists\footnote{Please note that the behavior of gradient and occlusion attribution techniques under feature heterogeneity (i.e. binary vs numerical), as well as sparsity, has not been extensively investigated and the results presented here could be affected by those factors.}, apart from Serum urea nitrogen and Serum creatinine. While it is reassuring to see these features in both rankings given their direct relationship to the definition of the AKI label \citep{Khwaja2012}, they do not represent the `cause' of the adverse event, and are hence not `actionable' from a clinical perspective. Interestingly, the ranking for gradients highlights an antibiotic (Nafcillin), used in our definition of the `Antibiotics' concept.

\begin{table*}[ht]
\begin{small}
\begin{tabular}{lcp{4.5cm}lcp{4.5cm}}
\multicolumn{3}{c}{Occlusion} & \multicolumn{3}{c}{Gradients} \\
\cmidrule(lr){1-3}\cmidrule(lr){4-6}
Feature type & Score & Feature Name & Feature type & Score & Feature Name\\
\cmidrule(lr){1-3}\cmidrule(lr){4-6}
Vitals, presence	&	0.0193	&	Bair hugger	&	Prescription, presence	&	0.0148	&	Nephrocaps	\\
Vitals, presence	&	0.0177	&	Nitric Oxide PPM	&	Prescription, presence	&	0.0042	&	Sulfameth/Trimethoprim (IV)	\\
Vitals, presence	&	0.0176	&	Nitric tank pressure	&	Admission, domain	&	0.0038	&	-	\\
Labs, presence	&	0.0132	&	Serum urea nitrogen	&	Labs, value	&	0.0036	&	Serum creatinine	\\
Vitals, presence	&	0.0127	&	Dialysis Access Type: catheter	&	Prescription, presence	&	0.0035	&	\textbf{Nafcillin} (IV)	\\
Vitals, presence	&	0.0120	&	Dialysis Type: peritoneal	&	Vitals, presence	&	0.0034	&	System Integrity: clots present	\\
Vitals, presence	&	0.0116	&	Emesis [Appearance]	&	Prescription, presence	&	0.0033	&	Furosemide	\\
Vitals, presence	&	0.0114	&	GI Tube Place Method: gastric ph	&	Vitals, presence	&	0.0030	&	Heparin Dose (per hour)	\\
Labs, presence	&	0.0103	&	Serum creatinine	&	Prescription, presence	&	0.0030	&	Atropine Sulfate (IV)	\\
Vitals, presence	&	0.0098	&	Diagnosis/op: hyperglycemia	&	Labs, value	&	0.0026	&	tbili	\\
Vitals, presence	&	0.0093	&	Dialysis Access Site: abdomen	&	Prescription, presence	&	0.0025	&	NS Epidural Bag	\\
Labs, value	&	0.0087	&	Serum urea nitrogen	&	Prescription, presence	&	0.0023	&	Midodrine HCl	\\
Vitals, presence	&	0.0085	&	PA Catheter Waveform Appear: unable to wedge	&	Vitals, presence	&	0.0021	&	Impaired Skin: extremities, lo	\\
Vitals, presence	&	0.0080	&	Allergy 1: thiazides	&	Prescription, presence	&	0.0018	&	Heparin (Hemodialysis)	\\
Vitals, presence	&	0.0077	&	Micro-Neb Treatment: alb/neb	&	Prescription, presence	&	0.0017	&	Nitroglycerin (IV drip)	\\
\end{tabular}
\end{small}
\caption{Occlusion and gradient attribution scores, averaged across patients and time steps for global explanations.}
\label{tab:mimic_aki_occlusion_gradients}
\end{table*}

For the patient presented in Figure~\ref{fig:mimic_antibio_global}, the predicted risk at time of antibiotics is low, and the model does not include this feature in its top 10 according to occlusion. Gradients however display Vancomycin, as well as heart rate and respiratory rate in the top 10, features that can be related to microbial infections. We see in Figure~\ref{fig:mimic_antibio_global}c that the alignment with the model's prediction does not happen at the time of antibiotics, but closer to the AKI event, highlighting the fact that TCAV alignment scores are not acting as `feature detectors'.

\begin{table}[ht]
\begin{small}
\begin{tabular}{p{4cm}p{4cm}}
Occlusion & Gradients \\
\cmidrule(lr){1-1}\cmidrule(lr){2-2}
Urine Output, pres & Lorazepam, pres \\
Potassium Chloride, pres & Heart Rate, value\\
Syringe (Neonatal), value & \textbf{Vancomycin}, pres\\
Noninvasive Systolic, value & Safety Measures, family, pres\\
Normocarb, pres & Calcium Gluconate, value\\
Potassium Chloride, value & Lactulose, pres\\
Docusate Sodium, pres & Respiratory Rate, value\\
Sodium Fluoride, pres & Soln., pres\\
Sodium Fluoride, value & D5W, pres\\
Urine Output, value & Senna, pres\\
\end{tabular}
\end{small}
\caption{Occlusion and gradient attribution scores, at time of antibiotics for the patient presented in Figure\ref{fig:mimic_antibio_global}.}
\label{tab:mimic_aki_occlusion_gradients_local}
\vskip -0.4cm
\end{table}
 
\section{Discussion and future work}
In this work, we explore the use of TCAV for RNNs, by defining concepts as trajectories over time. Across datasets and concepts, we notice that CAV$_{t_{start}:t_{end}}$ consistently leads to better generalization across time points. This result suggests that a majority of time steps in the time windows selected were relevant for the considered concepts. We note that extending this window or increasing the variability of the signals within this window might lead to different results. Nevertheless, we show that this approach provides meaningful CAVs and that both $tCA$ and CS scores are consistent with our expectations on the synthetic data. We observe that $tCA$ computed as a temporal ``derivative'' saturates if the concept's presence or absence does not vary over the time window $[t-dt,t]$. Similarly, CS highlights transitions in the model's predictions when taking the gradients w.r.t. the sigmoid of the logits, as the sigmoid saturates when the prediction is further away from the decision boundary. This could suggest a better use case of $tCA$ and CS in alert-based settings, where predictions/explanations are provided at specific time points, e.g. when the predicted risk passes the decision threshold.
% On the other hand, $tCA$ without lag and CS computed based on the gradients w.r.t. the logits could be used in continuous settings.

This work focuses on explaining predictions from RNNs and is hence bound by the model architecture. Recent works have however investigated other architectures for EHR predictions, including transformers \citep{Song2017} and point-wise convolutions \citep{Rocheteau2020}. While RNNs are adapted to the large number of features considered in the present model (tens of thousands compared to a couple of hundreds or less in newer architectures), future work could investigate how the approach developed here could be applied to other architectures.

While our approach factors in temporality in the construction of CAVs and $tCA$, CS would need to be extended to be able to account for how differences in model predictions relate to changes in the presence/absence of a concept \textit{over time}.  In this regard, a potential direction of work would be to refer to Temporal Integrated Gradients \citep{Hardt2019}. Such a method based on integrated gradients ~\citep{Sundararajan2017} would also enable the use of the proposed approach for local explanations, as integrated gradients estimate the difference between the obtained prediction and a ``neutral decision''.

One limitation of TCAV arises from the difficulty of defining a concept through examples from real-world EHR data. While toy datasets or ImageNet applications seem intuitive, healthcare data can be difficult to separate into clinical concepts. In the present work, we computed relative CAVs by selecting concept and control examples using simple rules, typically based on one (`sex') or an ensemble of features. Our results suggest that the proposed approach can capture correlated signals and trajectories based on this proxy definition, and do not act as `feature detectors'. Based on their definition, concepts can be defined to represent `actionable' clinical concepts and can investigate signals that would not be directly represented as features in the model training set (e.g. gender). This is in contrast to feature-based attributions, that are tied to the features present in the samples used to compute attributions. However, concepts, while designed to be human-understandable, will typically encompass multiple features, as well as potential changes in patterns of those features across time. Therefore, feature-based attributions could be used in conjunction with concept-based attributions to understand which specific features could affect the model's alignment with a concept, e.g. which physiological features have been affected by the microbial infection as treated by the antibiotics.

We further matched control examples for a number of criteria. We note that other matching criteria and methods could be used, e.g. propensity score matching or other optimal transport techniques. While we reported promising results on multiple concepts using these CAVs, a danger is to miss confounding factors that then lead to a significant CAV. A future direction could be to generate counterfactuals, as in ~\citep{Goyal2019, Pfohl2019, Singla2019}. Given the dimensionality of the data, training and evaluating such a counterfactual generative model however remains challenging. Another risk lies in potential confirmation bias during the process of building the CAV and estimating $tCA$ and CS scores, as the user might be tuning the CAV building until a concept surfaces. It could hence be desirable to know how much signal is covered by a set of concepts, as proposed in ~\citep{Yeh2019}. On the other hand, the clinical user might want to define a limited set of ``actionable'' concepts, e.g. ``dehydration'' or ``nephrotoxicity'', for which a clinical action could prevent the predicted outcome. This would alleviate the concerns around building the ``complete'' set of concepts, and provide a path to action, especially in the case of local explanations. We also note that defining concepts requires the involvement of clinicians. We believe that this is a strength of the method rather than a weakness, as it allows clinicians to define what ``actionable'' or ``trustworthy'' mean in the selected use case, leading to increased transparency in the machine learning development pipeline.

Finally, we evaluate the proposed approach empirically, based on the `ground truth' present in the synthetic data, as well as on a clinical benchmark. Our concepts and results for MIMIC were assessed by a clinician. Future work should however assess the clinical relevance and utility of TCAV for EHR  more rigorously, using human and task grounded evaluations, as suggested in ~\citep{Doshi-Velez2018}. In particular, it would be useful to investigate whether concept-based explanations can help guide clinical actions taken in response to a prediction, and, ultimately, whether these explanations improve outcomes for patients.

\section*{Software and Data}
The Python and TensorFlow code to generate the synthetic dataset, models, and compute CS and $tCA$ is available on Github at \url{https://github.com/google/ehr-predictions/tree/master/tcav-for-ehr}. The de-identified EHR data is available based on a user agreement at \url{https://physionet.org/content/mimiciii/1.4/}.

\begin{acks}
We thank Been Kim and Yash Goyal for discussions and for sharing code.
\end{acks}

\bibliographystyle{ACM-Reference-Format}
\bibliography{refs}

\appendix

\section{Synthetic data}
\subsection{Data generation}\label{apd:synth_data}
\begin{figure*}[!ht]
\begin{center}
\centerline{\includegraphics[width=.7\textwidth]{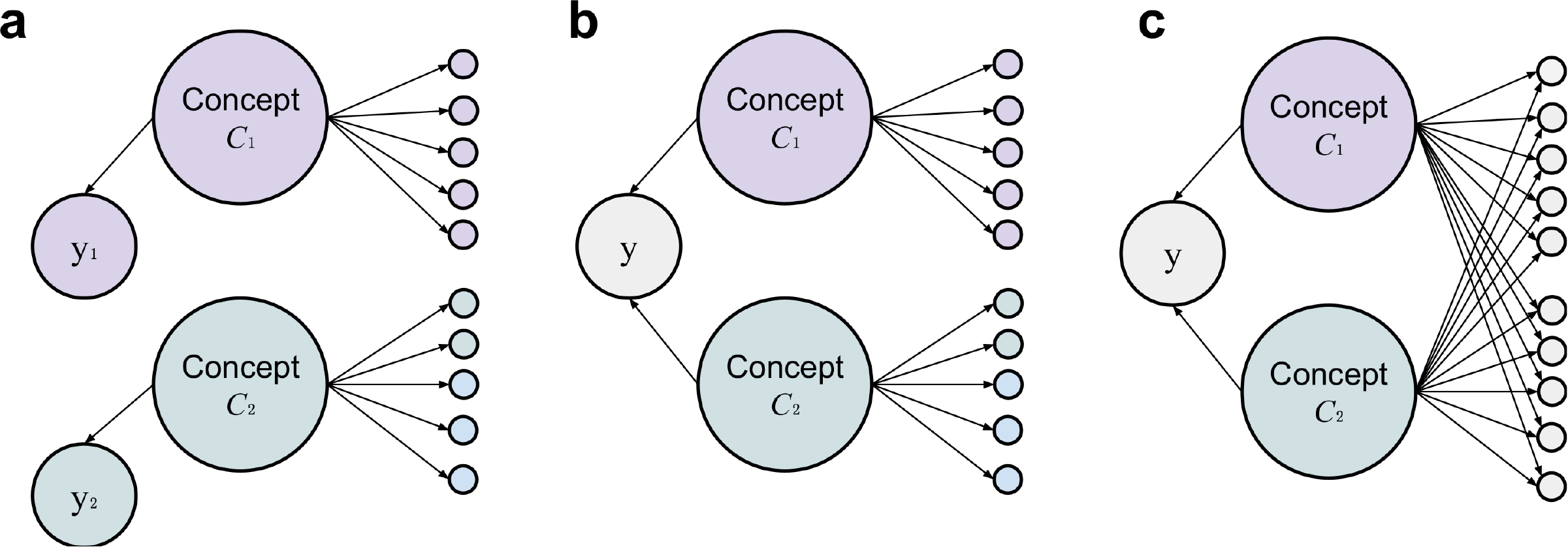}}
\caption{Illustration of the causal graph and potential uses. \textbf{a} Concepts sampled independently, each affecting non-overlapping sets of features and different labels. \textbf{b} The concepts jointly affect the label (e.g. using AND or OR). \textbf{c} The concepts jointly affect the label and all features.}
\label{fig:synthetic_data_causal_graph}
\end{center}
\end{figure*}

The dataset has been designed to allow for multiple different scenarios (Figure~\ref{fig:synthetic_data_causal_graph}). For simplicity, the data generation process used throughout the main text relies on scenario \textbf{a} of Figure~\ref{fig:synthetic_data_causal_graph}.

The data is sampled according to Algorithm~\ref{alg:data_generation}.

\begin{algorithm}[ht]
\caption{Generating one synthetic time series $\vec{x}$, given $c$ concepts, $T$ timesteps and $D$ features. $\delta_C$ is a binary variable defining whether a concept $C$ is `present' or `absent'. $\lambda_{d, C}$ is a binary variable that defines whether concept $C$ manifests in feature $d$.}
\label{alg:data_generation}
\ForAll{$C \in {C_1, C_2,\dots, C_{c}}$}{
Sample $t_{start} \sim \texttt{Uniform}(\{1,...,T\})$\;
Sample $\delta_C \sim \texttt{Bernoulli}(0.5)$\;
Sample the label $y_{C,t_{start}:T} \sim \texttt{Bernoulli}(p(y|C))$\;
\ForAll{$d \in {1,\dots,D}$}{
Sample $\lambda_{d, C} \sim \texttt{Bernoulli}(p(d|C))$\;
\If{$\vec{x}_{d}$ is numerical}{
$\vec{x}_{d} \sim \mathcal{N}(0, 0.5)$ \; 
\lIf{$\lambda_{d, C}$}{$\vec{x}_{d, t_{start} : T} \sim \mathcal{N}(0, 0.5) + \sin{(\vec{x}_{d, t_{start} : T})} $}
}
\If{$\vec{x}_{d}$ is binary}{
$\vec{x}_{d} \sim \texttt{Bernoulli}(0.5)$\;
\lIf{$\lambda_{d,C}$}{$\vec{x}_{d, t_{start} : T} \sim \texttt{Bernoulli}(0.95)$}}
}
}
\end{algorithm}

\subsection{Concept-based explanations}\label{apd:synthetic_results}

\subsubsection{CAV building}\label{apd:synth_CAVs}
We display the balanced accuracy and ROC AUC of the linear CAV classifiers evaluated on held-out samples, as well as on further time series (including time points outside the $[t_{start}:t_{end}]$ window) for each building strategy in tables Table~\ref{tab:synthetic_CAVs_t0_t1} (CAV$_{t_{start}:t_{end}}$), Table~\ref{tab:synthetic_CAVs_t1} (CAV$_{t_{end}}$) and Table~\ref{tab:synthetic_CAVs_t1_t0} (CAV$_{t_{end}-t_{start}}$).

\begin{table*}[!ht]
\begin{small}
\begin{tabular}{ccccc}
\toprule
\bfseries Concept & \bfseries Layer & \bfseries Accuracy [\%] & \bfseries ROC AUC [\%] & \bfseries Accuracy on test [\%]\\
\midrule
$C_1$ & 0 & 88.97 & 95.89 & 87.72\\
$C_1$ & 1 & 89.08 & 96.12 & 89.44\\
$C_1$ & 2 & 91.55 & 97.33 & 90.52\\
$C_2$ & 0 & 91.78 & 97.12 & 86.67\\
$C_2$ & 1 & 91.18 & 97.11 & 89.15\\
$C_2$ & 2 & 92.93 & 97.51 & 90.65\\
\bottomrule
\end{tabular}
\end{small}
\caption{Performance of CAVs on held-out and test samples, CAV$_{t_{start}:t_{end}}$ building strategy.}
\label{tab:synthetic_CAVs_t0_t1}
\end{table*}

\begin{table*}[!ht]
\begin{small}
\begin{tabular}{ccccc}
\toprule
\bfseries Concept & \bfseries Layer & \bfseries Accuracy [\%] & \bfseries ROC AUC [\%] & \bfseries Accuracy on test [\%]\\
\midrule
$C_1$ & 0 & 93.51 & 98.27 & 85.08\\
$C_1$ & 1 & 94.34 & 99.55 & 90.31\\
$C_1$ & 2 & 95.46 & 99.53 & 92.51\\
$C_2$ & 0 & 85.29 & 93.41 & 83.40\\
$C_2$ & 1 & 89.90 & 96.24 & 91.02\\
$C_2$ & 2 & 92.15 & 97.11 & 91.22\\
\bottomrule
\end{tabular}
\end{small}
\caption{Performance of CAVs on held-out and test samples, CAV$_{t_{end}}$ building strategy.}
\label{tab:synthetic_CAVs_t1}
\end{table*}

\begin{table*}[!ht]
\begin{small}
\begin{tabular}{ccccc}
\toprule
\bfseries Concept & \bfseries Layer & \bfseries Accuracy [\%] & \bfseries ROC AUC [\%] & \bfseries Accuracy on test [\%]\\
\midrule
$C_1$ & 0 & 71.37 & 80.25 &  81.10\\
$C_1$ & 1 & \textit{73.04*} & \textit{83.37*} & 82.96\\
$C_1$ & 2 & 76.40 & 86.54 & 83.15\\
$C_2$ & 0 & 69.96 & 79.14 & 76.35\\
$C_2$ & 1 & \textit{72.31*} & \textit{79.29*} & 84.35\\
$C_2$ & 2 & \textit{74.21*} & \textit{83.18*} & 82.37\\
\bottomrule
\end{tabular}
\end{small}
\caption{Performance of CAVs on held-out and test samples, CAV$_{t_{end}-t_{start}}$ building strategy. Non-significant results after correction for multiple comparisons are emphasized.}
\label{tab:synthetic_CAVs_t1_t0}
\end{table*}

\begin{figure}[!ht]
\includegraphics[width=\columnwidth, left]{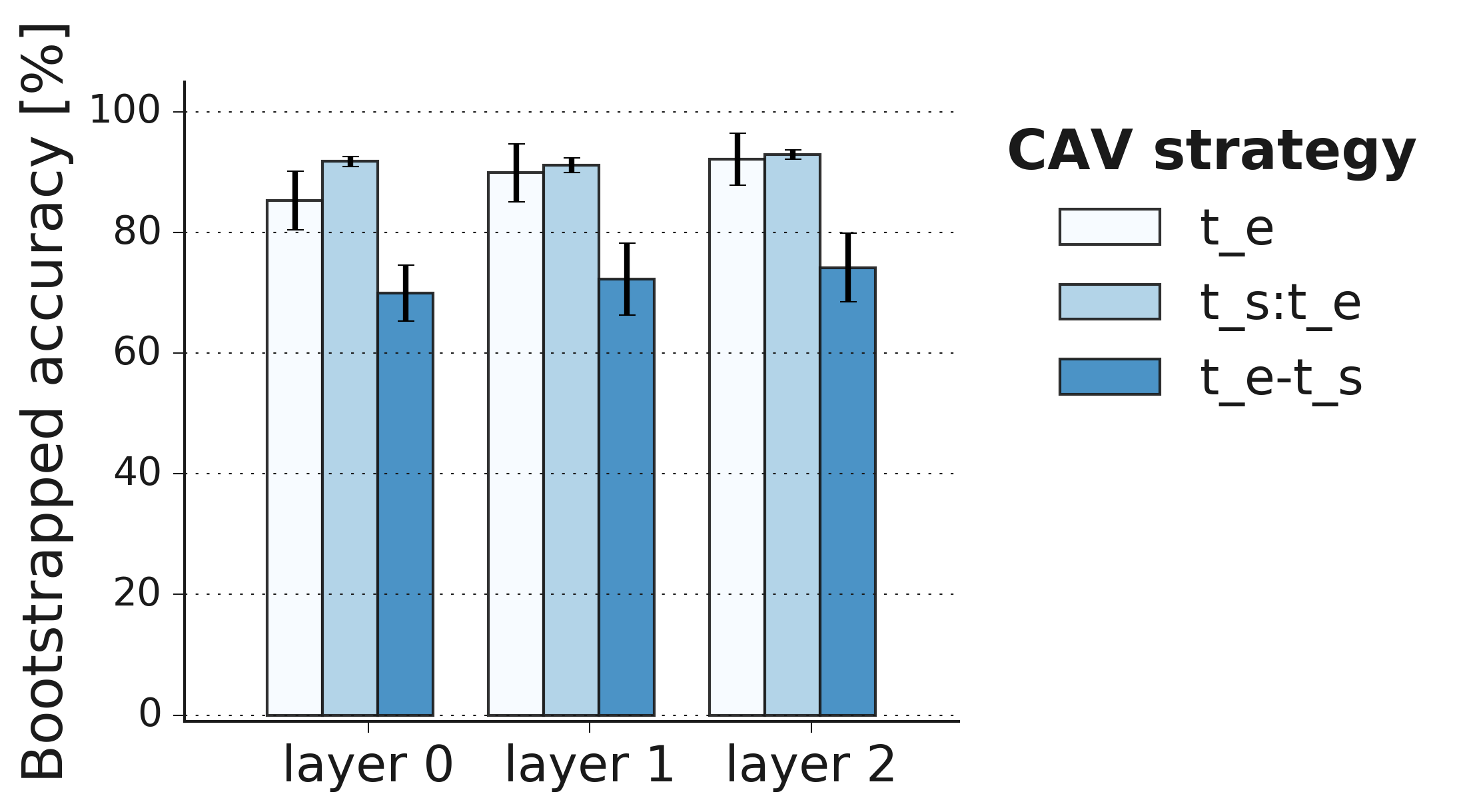} \\
\includegraphics[width=0.68\columnwidth, left]{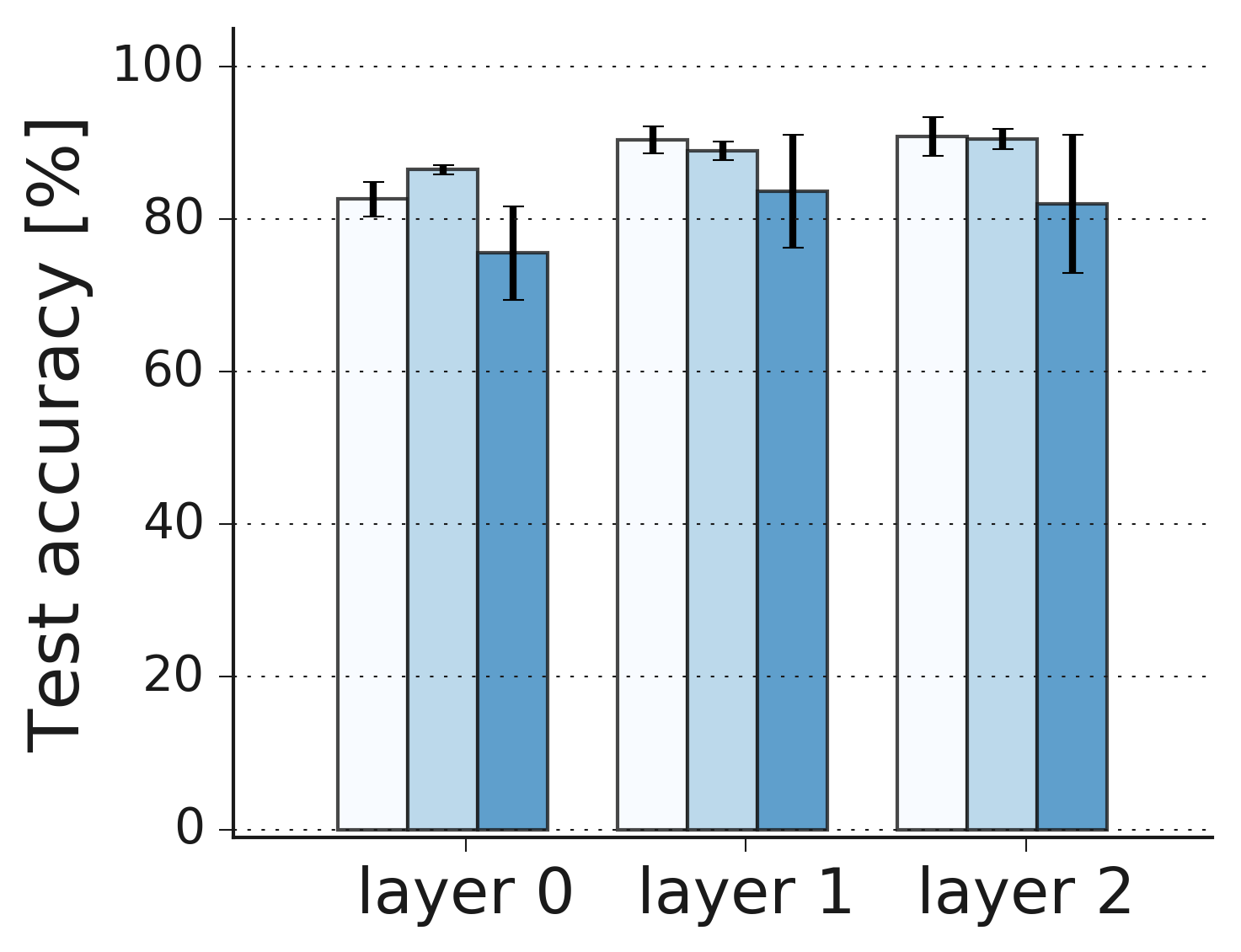}
\caption{\textbf{Synthetic CAV results.} Accuracy of the $C_2$ CAV in bootstrap (top) and test (bottom) evaluations for CAV$_{t_{end}}$, CAV$_{t_{start}:t_{end}}$ and CAV$_{t_{end}-t_{start}}$, in \%.}
\label{fig:synth_cavs_C2}
\end{figure}

Figure~\ref{fig:synth_cavs_C2} illustrates the balanced accuracy (in \%) for the linear classifier defining $C_2$.

\subsubsection{Influence of the CAV}
\begin{figure*}[!ht]
\includegraphics[width=\linewidth]{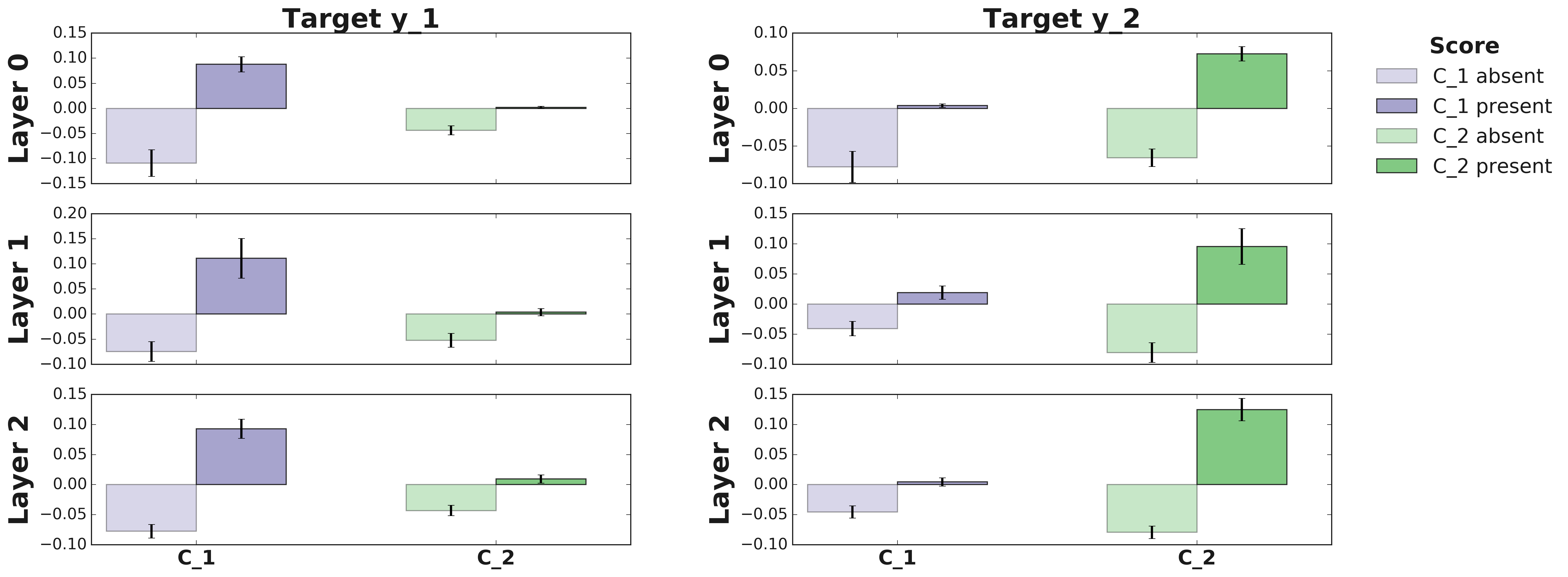}
\caption{CS scores as averaged across all time steps and bootstraps (mean$\pm$std) for each target and layer when a concept ($C_1$ purple, $C_2$ green) is absent (light shade) or present (dark shade).}
\label{apd:CS_synth_bars}
\end{figure*}

We display the global CS scores for each layer and target when the concepts are present or absent in Figure~\ref{apd:CS_synth_bars}.

\begin{figure*}[!ht]
\includegraphics[width=0.8\linewidth]{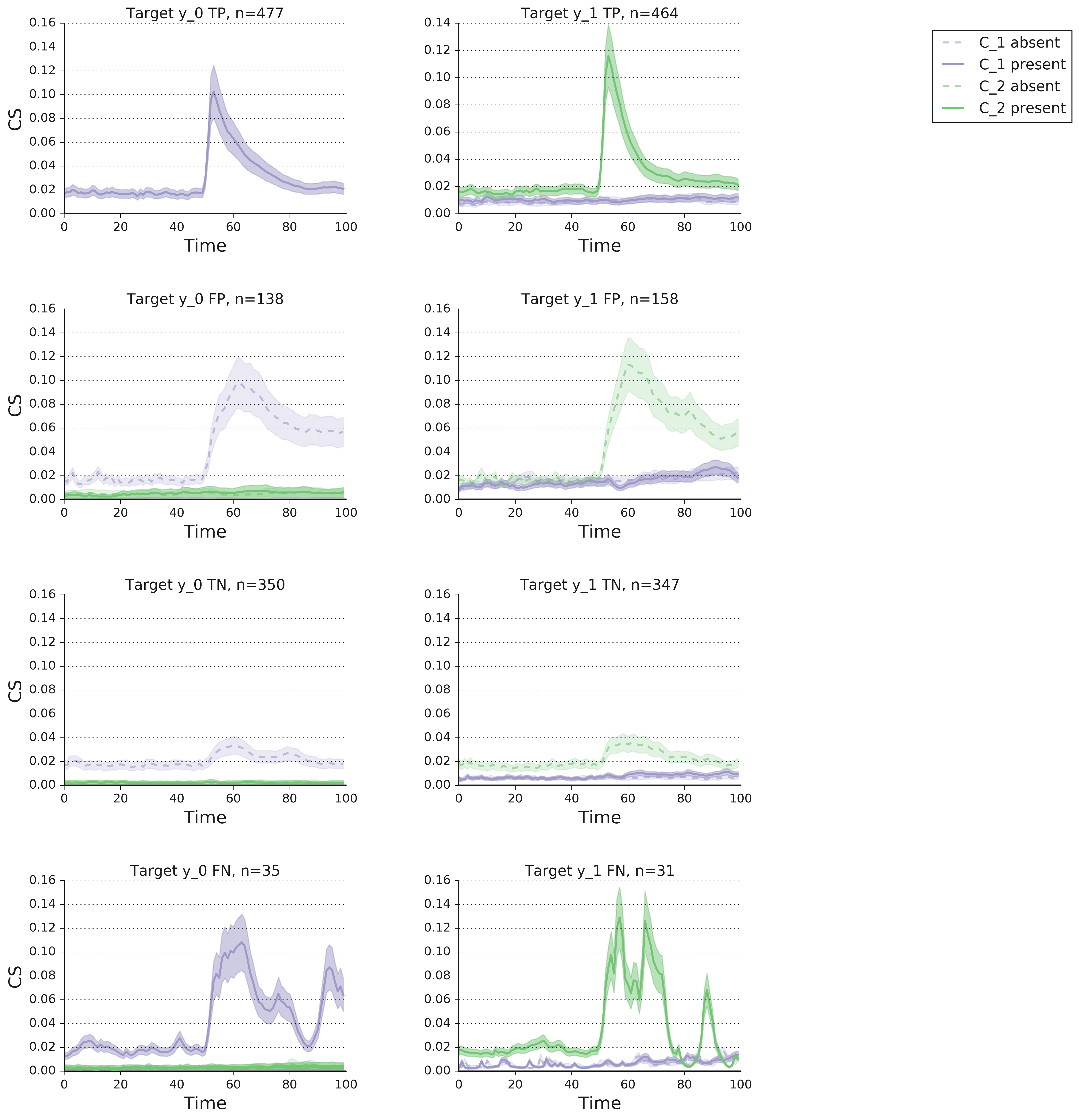}
\caption{CS scores at layer 2 as averaged across all time series (mean$\pm$std) for each target when a concept ($C_1$ green, $C_2$ purple) is absent (light shade, dotted) or present (dark shade), for correct (TP, TN) and incorrect (FP, FN) predictions. $n$ represents the number of trajectories considered for averaging.}
\label{apd:CS_synth_TP_FP}
\end{figure*}

Figure~\ref{apd:CS_synth_TP_FP} displays global CS scores per time point across the different categories of positive and negative predictions that the model makes.

\section{Application to MIMIC}

\subsection{Data and modeling}
\subsubsection{Data representation}
\label{apd:sup_EHR_data}
 The de-identified EHR data were mapped to the Fast Healthcare Interoperability Resource (FHIR) specification as described in~\cite{Rajkomar2018}.  From these records, we extracted the following feature categories: admission metadata, observations, procedures, diagnoses, laboratory tests, medication orders and vital signs. No free-text clinical notes were used. These records were organised into a sequential patient representation, where all data for a single patient were ordered by time, subdivided into admissions and then bucketed into 1-hour time steps, taking the median of any features present multiple times per time step. Continuous features were associated with an additional binary `presence' variable to indicate whether the feature was present or absent at that time step. No feature imputation was applied. Categorical features were one-hot encoded. All features were trimmed at the 1 and 99 percentiles, and standardized.

\subsubsection{Model architecture and training}
We use the model architecture described in \cite{Tomasev2019} to predict acute kidney injury in a range of future prediction windows ranging from 6h ahead to 72h. In particular, the model used in this analysis comprises 2 embedding layers of size 400 with residual connections and an RNN architecture using a GRU cell of 200 units per layer and 3 layers. We use a batch size of 128, a back propagation through time window of 128, and an initial learning rate of 0.001 that decays every 12,000 steps by a factor of 0.85. These parameters were selected via a grid search using the validation set. Learning is stopped after the earlier of 100,000 steps or AUPRC convergence on the task of predicting AKI 48 hours ahead of time. The model is evaluated on the test set using AUPRC: we observe AUPRC of 41.0\% for the task of predicting any severity of AKI in the next 24 hours, and AUPRC of 44.6\% for the task of predicting any severity of AKI in the next 48 hours.

\subsection{AKI concept}

\paragraph{CAV building}\label{apd:mimic_aki_cavs}

Table~\ref{tab:mimic_CAV_performance_aki2} displays the performance of the CAV linear classifier as evaluated on held-out samples from the bootstrap scheme. All building strategies lead to significant linear classifiers. When evaluated on time series from different patients, we however notice that performance decreases significantly for CAV$_{t_{end}}$ (accuracy $\sim$ 30\% for concept group time steps) and CAV$_{t_{end}-t_{start}}$ (accuracy $\sim$ 40\% for concept group time steps). This result suggests that those strategies, which focus most on $t_{end}$ might be overfitting to this specific time step.

\begin{table*}[ht]
\begin{small}
\begin{tabular}{lccccc}
\toprule
\bfseries  CAV building strategy & \bfseries Layer & \bfseries  n$_{concept}$ & \bfseries  n$_{control}$ & \bfseries  Accuracy [\%] & \bfseries  ROC AUC [\%] \\
\midrule
 & 0 & & & 87.03 & 94.60 \\
Last 12h (CAV$_{t_{start}:t_{end}}$)  & 1 & 317 & 306 & 80.18 &  90.03 \\
 & 2 & & & 75.64 &  87.92 \\
\hline
 & 0 & & & 88.53 &  95.03 \\
Last 24h (CAV$_{t_{start}:t_{end}}$)  & 1 & 599 & 572 & 82.04 &  89.67 \\
 & 2 & & & 76.15 &  86.96 \\
\hline
 & 0 & & & 91.17 &  96.62 \\
Time of AKI (CAV$_{t_{end}}$)  & 1 & 81 & 80 & 86.95 &  96.12 \\
 & 2 & & & 80.81 &  92.29 \\
\hline
 & 0 & & & 90.00 &  95.72 \\
AKI - 12h (CAV$_{t_{end}-t_{start}}$)  & 1 & 81 & 80 & 85.22 &  95.03 \\
 & 2 & & & 74.97 &  89.01 \\
\bottomrule
\end{tabular}
\end{small}
\caption{Performance of CAV classifier for the \textit{AKI} concept on MIMIC, averaged across 100 bootstrap resamples.}
\label{tab:mimic_CAV_performance_aki2}
\end{table*}

\subsection{NSAIDs concept}

\subsubsection{NSAIDs definition}\label{apd:nsaids}
Our computable definition for NSAID exposure was based on the US Food and Drug Administration (FDA) Established Pharmacological Class (EPC) grouping, with two additional COX-2 selective NSAIDs that were used historically but are no longer on this list (rofecoxib and valdecoxib). The complete list of drugs included is presented in table \ref{nephrotoxic_drugs}. Previous studies on NSAID-mediated nephrotoxicity have either excluded aspirin and acetaminophen due to their widespread use at low doses~\citep{Plantinga2011} or only used high-dose aspirin~\citep{lafrance}; however aspirin is excluded here to avoid the need for dose calculations in this preliminary work. 

\begin{table}[!t]
\begin{center}
\begin{small}
\begin{sc}
\begin{tabular}{@{}ll@{}}
\toprule
Medication name &  \\ \midrule
bromfenac & ketorolac \\ 
celecoxib  & mefenamic acid \\
diclofenac & meloxicam \\
diflunisal & nabumetone \\
etodolac & naproxen \\
fenoprofen & oxaprozin \\
flurbiprofen & piroxicam \\
ibuprofen & rofecoxib \\
indomethacin & sulindac \\
ketoprofen & valdecoxib \\
\bottomrule
\end{tabular}
\end{sc}
\end{small}
\end{center}
\caption{List of NSAID medications included in the \textit{nephrotoxicity} concept.}
\label{nephrotoxic_drugs}
\end{table}

\subsubsection{CAV building}\label{apd:mimic_nsaids_cavs}

Table~\ref{tab:mimic_CAV_performance_nsaids_before_aki} displays the performance of the CAV linear classifier as evaluated on held-out samples from the bootstrap scheme. The results suggest that the difference between the concept and control groups is more subtle, which is not unexpected. As for the AKI concept, the CAV$_{t_{start}:t_{end}}$ seems to generalize better to other time series and time steps (consistent accuracy for both control and concept time series, $\sim 55 - 65\%$, compared to unstable results for CAV$_{t_{end}-t_{start}}$).

\begin{table*}[ht]
\begin{small}
\begin{tabular}{lccccc}
\toprule
\bfseries  CAV building strategy & \bfseries Layer & \bfseries  n$_{concept}$ & \bfseries  n$_{control}$ & \bfseries  Accuracy [\%] & \bfseries  ROC AUC [\%] \\
\midrule
 & 0 & & & 79.63 &  88.02 \\
Last 24h (CAV$_{t_{start}:t_{end}}$)  & 1 & 480 & 472 & 66.47 &  75.62 \\
 & 2 & & & 60.34 &  70.59 \\
\hline
 & 0 & & & 82.93 &  91.28 \\
NSAIDs to AKI (CAV$_{t_{start}:t_{end}}$)  & 1 & 503 & 470 & 70.70 &  81.49 \\
 & 2 & & & 65.70 &  78.27 \\
\hline
 & 0 & & & 75.15 &  83.71 \\
AKI - NSAIDs (CAV$_{t_{end}-t_{start}}$)  & 1 & 65 & 65 & 72.15 & 82.55 \\
 & 2 & & & 67.92 &  84.43 \\
\bottomrule
\end{tabular}
\end{small}
\caption{Performance of CAV classifier for the \textit{NSAIDs} concept on MIMIC, averaged across 100 bootstrap resamples.}
\label{tab:mimic_CAV_performance_nsaids_before_aki}
\end{table*}

\subsubsection{Local examples:}
\label{apd:mimic_nsaids_local}
\begin{figure*}[!ht]
\begin{center}
\includegraphics[width=0.9\textwidth]{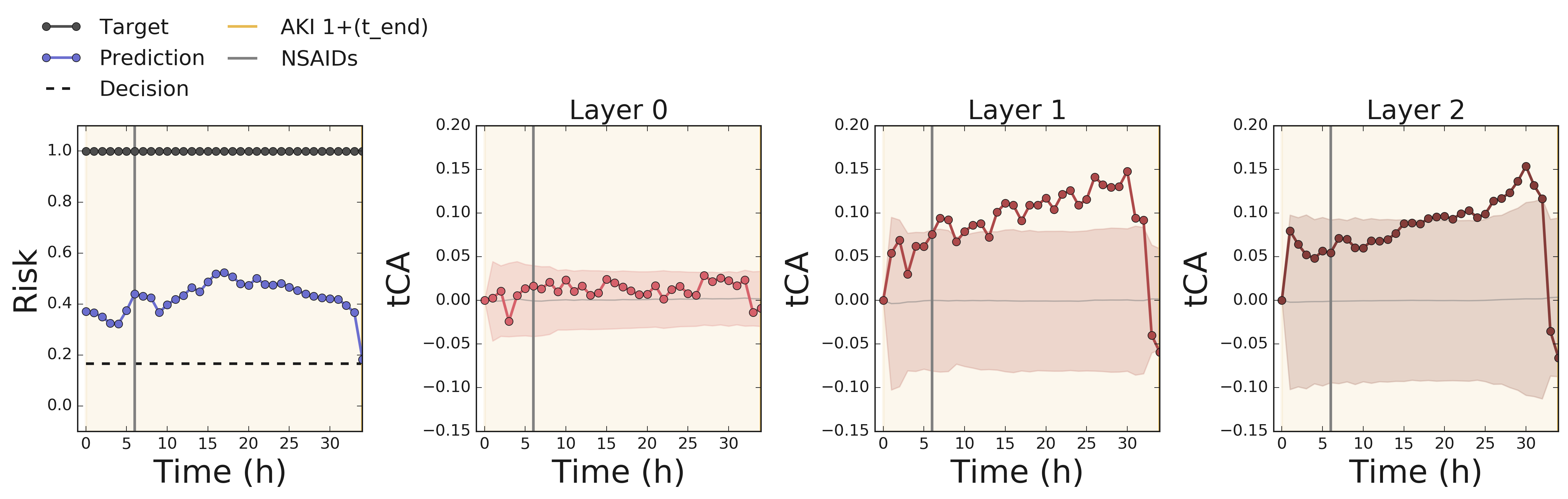} \\
\includegraphics[width=0.9\textwidth]{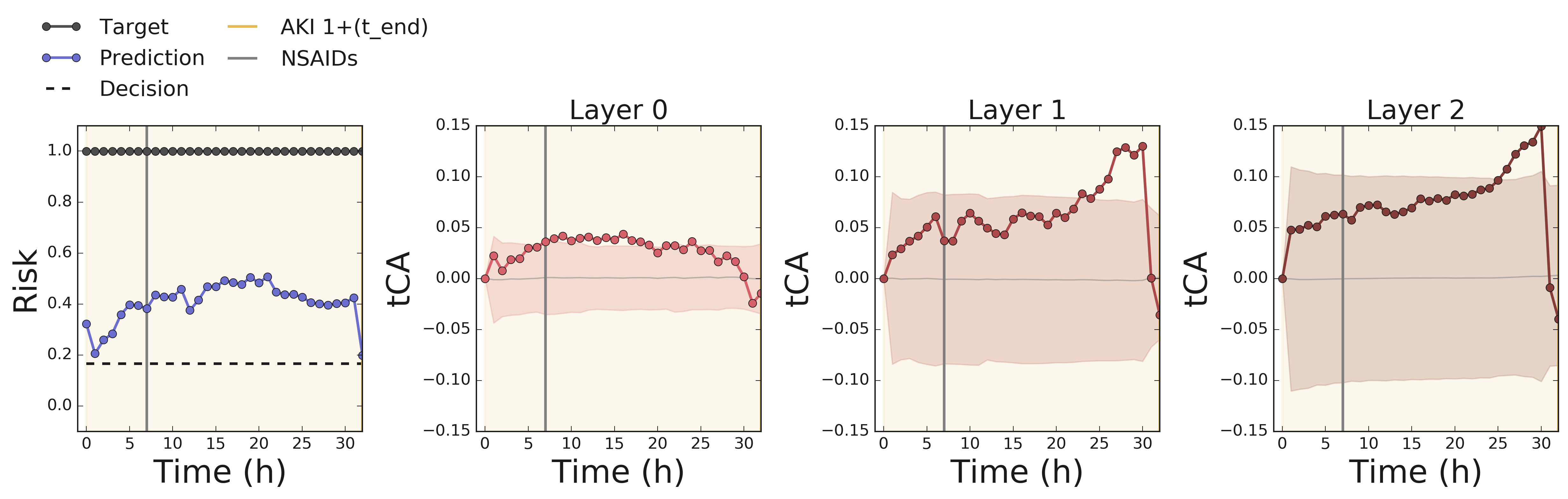} \\
\caption{\textbf{Local MIMIC results, NSAIDs concept, true positives.} Single patient timeseries, displaying the label (AKI 1+ within 48h) and model's output, as well as $tCA$ for each layer, and its null hypothesis (shaded). The yellow shaded area represents the prediction horizon of the model, i.e. 48 hours. The administration of NSAIDs is displayed by a grey vertical line. The first 2 rows display increases in $tCA$ in true positive predictions, the middle 2 rows display no or negative alignment for negative predictions and the last row displays an increase in alignment at time of NSAIDs administration in a false positive prediction.}
\label{fig:mimic_nsaids_local}
\end{center}
\end{figure*}

Figure~\ref{fig:mimic_nsaids_local} displays further local examples for the NSAIDs concept.

\subsection{Antibiotics concept}

\subsubsection{Antibiotics definition}\label{apd:antibiotics}
Similarly to NSAIDs, our computable definition for antibiotics exposure was based on the US Food and Drug Administration (FDA) Established Pharmacological Class (EPC) grouping. For this definition, we focus on antimicrobial agents (Table~\ref{apd_tab:antibiotics}). 

\begin{table*}[!ht]
\begin{center}
\begin{small}
\begin{sc}
\begin{tabular}{@{}ll@{}}
\toprule
Medication name &  \\ \midrule
Amikacin & Gentamicin\\
Ampicillin & Imipenem/Cilastatin\\
Ampicillin/Sulbactam (Unasyn) & Isoniazid\\
Atovaquone & Keflex\\
Azithromycin & Levofloxacin\\
Aztreonam & Linezolid\\
Bactrim (SMX/TMP) & Meropenem\\
Cefazolin & Metronidazole\\
Cefepime & Moxifloxacin\\
Ceftazidime & Nafcillin\\
Ceftriaxone & Oxacillin\\
Ciprofloxacin & Penicillin G potassium\\
Clindamycin & Piperacillin\\
Colistin & Piperacillin/Tazobactam (Zosyn)\\
Daptomycin & Pyrazinamide\\
Doxycycline & Rifampin\\
Erythromycin & Tobramycin\\
Ethambutol & Vancomycin\\
\bottomrule
\end{tabular}
\end{sc}
\end{small}
\end{center}
\caption{List of antimicrobial medications included in the \textit{antibiotics} concept.}
\label{apd_tab:antibiotics}
\end{table*}

\subsubsection{Local examples}
\label{apd:mimic_antibio_results}
\begin{figure*}[!ht]
\begin{center}
\includegraphics[width=0.9\textwidth]{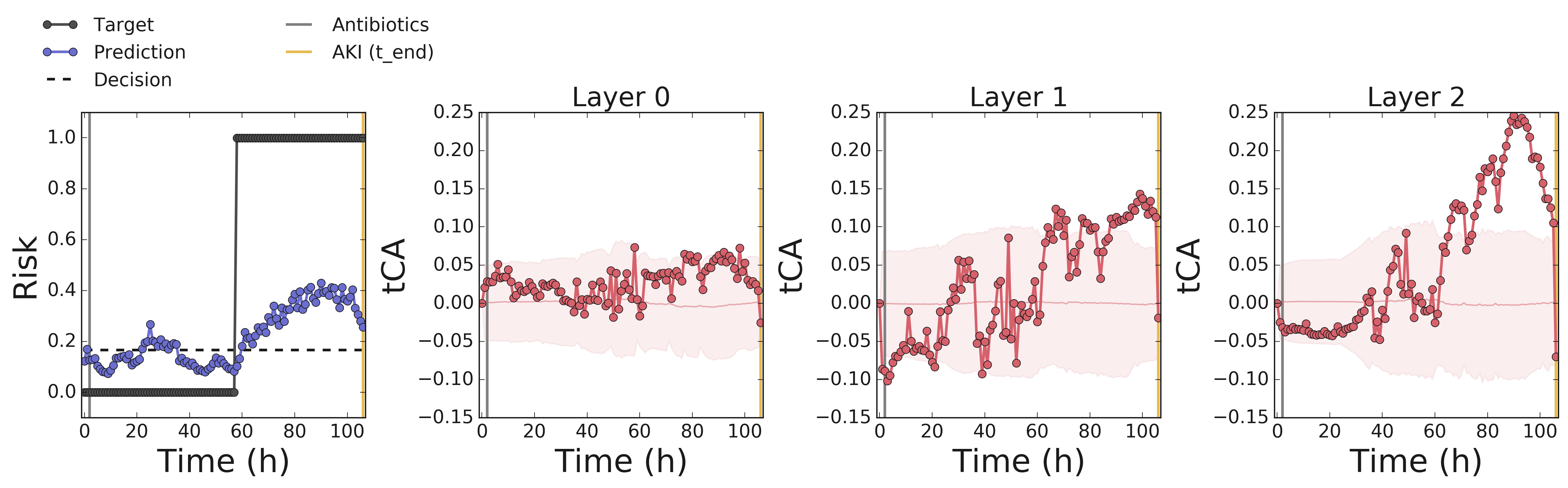} \\
\includegraphics[width=0.9\textwidth]{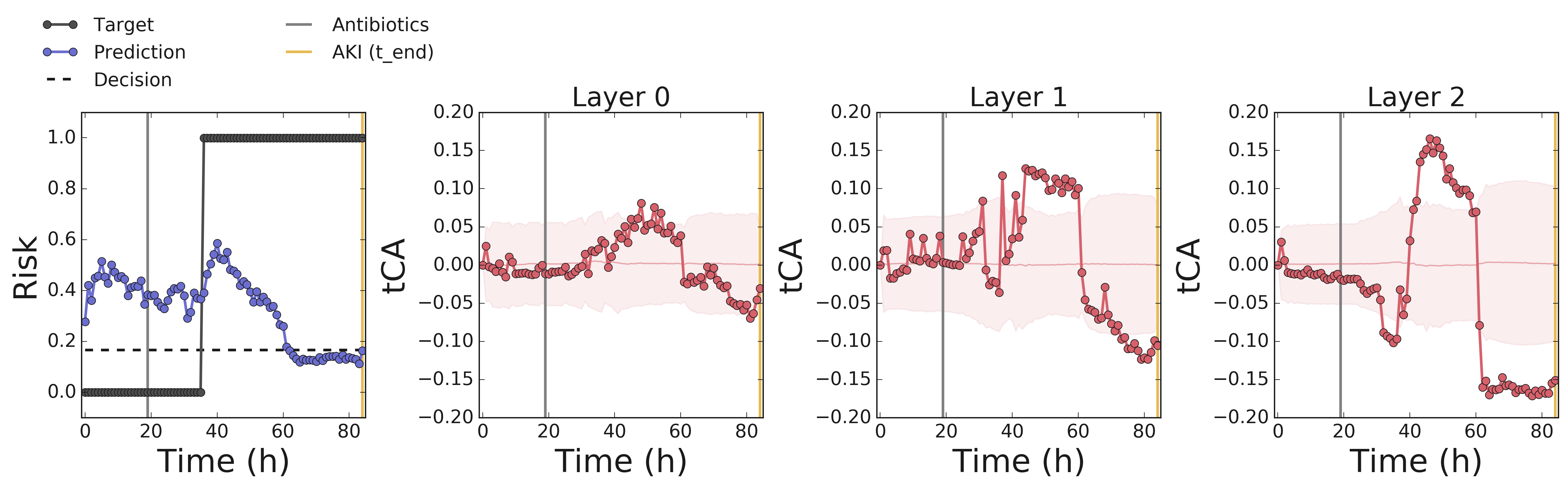} \\
\includegraphics[width=0.9\textwidth]{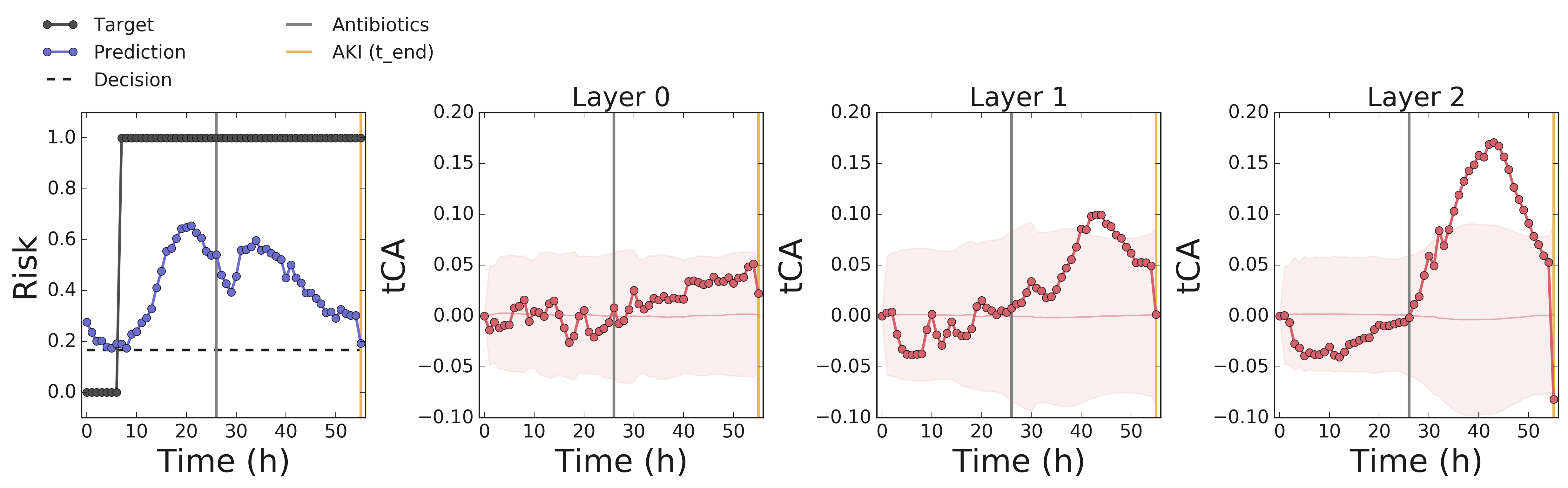} \\
\includegraphics[width=0.9\textwidth]{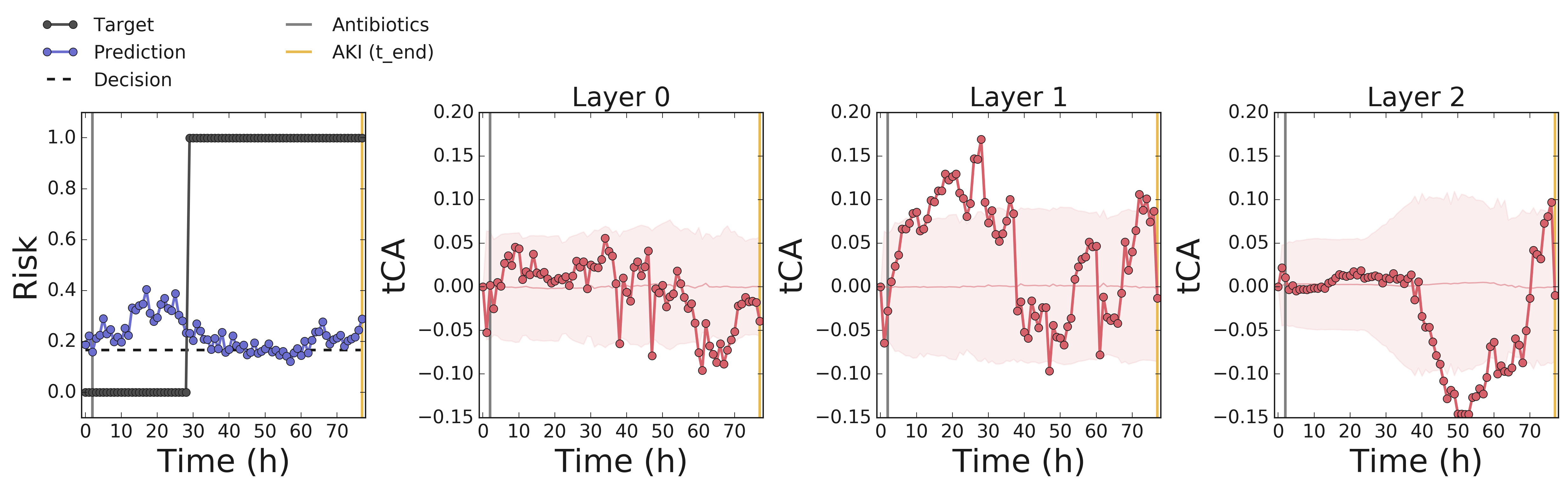}
\caption{\textbf{MIMIC results, Antibiotics concept.} Local patient trajectory with time of antibiotics displayed as a vertical grey line.}
\label{fig:mimic_antibio_local}
\end{center}
\end{figure*}

Figure~\ref{fig:mimic_antibio_local} displays further local results for the antibiotics concept. Interestingly, we see that layers do not necessarily all show an alignment between the concept and the model's activations. Further work will investigate the `sensitivity' of the method by assessing each patient's records and notes individually based on clinical input.

\subsection{Sex concept}
\label{apd:mimic_sex_concept}

\subsubsection{CAV building}
Table~\ref{tab:mimic_CAV_performance_sex_aki1p} displays the balanced accuracy and ROC AUC scores for the considered CAV building strategies, with their corresponding p-value as estimated by non-parametric permutation testing.

\begin{table*}[ht]
\begin{small}
\begin{tabular}{lccccccc}
\toprule
\bfseries  CAV building strategy & \bfseries Layer & \bfseries  n$_{concept}$ & \bfseries  n$_{control}$ & \bfseries  Accuracy pvalue & \bfseries Accuracy [\%] & \bfseries ROC AUC pvalue & \bfseries  ROC AUC [\%] \\
\midrule
 & 0 & & & 0.16 & 51.95 & 0.21 & 52.32 \\
Beginning of admission  (CAV$_{t_{start}:t_{end}}$) & 1 & 500 & 500 & 0.41 & 50.27 & 0.37 & 50.85 \\
 & 2 & & & 0.77 & 48.88 & 0.79 & 48.08 \\
\hline
 & 0 & & & 0.60 & 49.50 & 0.56 & 49.72 \\
Last 12h (CAV$_{t_{start}:t_{end}}$) & 1 & 500 & 500 & 0.47 & 50.07 & 0.53 & 49.91 \\
 & 2 & & & 0.64 & 49.43 & 0.63 & 49.21 \\
\hline
 & 0 & & & 0.61 & 49.45 & 0.63 & 49.09 \\
Last 24h (CAV$_{t_{start}:t_{end}}$) & 1 & 500 & 500 & 0.40 & 50.34 & 0.38 & 50.78 \\
 & 2 & & & 0.36 & 50.59 & 0.32 & 51.28 \\
\bottomrule
\end{tabular}
\end{small}
\caption{Performance of CAV classifier for the \textit{Sex} concept on MIMIC, averaged across 100 bootstrap resamples.}
\label{tab:mimic_CAV_performance_sex_aki1p}
\end{table*}

\subsection{Feature-based attributions}
\label{apd:feature_based_methods}

\subsubsection{Gradients}
The absolute values of the gradients of a network provide an estimation of how much the variables need to be perturbed to provide a change in output ~\cite{Ancona2018}. For time series, they are computed as ~\cite{Hardt2019}:
\begin{equation}
    g^{t_1}_{i, t} = \frac{\partial F_{t_1}(\textbf{x})}{\partial x_{i, t}}
\end{equation}
Where $g^{t_1}_{i,t}$ represents the attribution for variable $i$ at time $t$ given the risk at $t_1$. We estimate gradients at $t=t_1$ for each time step, i.e. computing the instantaneous gradients. To obtain global attributions, we average the absolute value of gradients (normalized at each time step) across time steps and patients.

\subsubsection{Occlusion}
Occlusion \cite{Zeiler2014} evaluates the change in model output if a variable is occluded from the example(s). For each feature $i$ at time step $t$, we compute:
\begin{equation}
    o_{i,t} = F_t(\vec{x}) - F_t(\vec{x} | x_{i,t} = b)
\end{equation}

where $o_{i,t}$ represents the occlusion score for variable $i$ at time $t$ and $b$ represents a chosen baseline. For binary variables that are `present' at a specific time step, we occlude their presence by setting it to 0. For numerical values, we replace the value by 0. This corresponds to replacing by the population mean, given the normalization scheme used in preprocessing.

\end{document}